%% file: main_pdf_not_eps.tex
\documentclass[Afour,sageh,times]{sagej}

\usepackage{moreverb,url}
\usepackage{tabularx}

\usepackage[all]{nowidow}

\NewDocumentCommand{\rot}{m}{\rlap{\hspace*{-1mm}\rotatebox{55}{#1}}}

\newcommand\BibTeX{{\rmfamily B\kern-.05em \textsc{i\kern-.025em b}\kern-.08em
T\kern-.1667em\lower.7ex\hbox{E}\kern-.125emX}}

\def\volumeyear{2025}

\input{preamble.tex}
\begin{document}

\widowpenalty10000
\clubpenalty10000

\runninghead{Boxan et al.}

\title{FoMo: A Multi-Season Dataset for Robot Navigation in Forêt Montmorency}

\author{Matěj Boxan\affilnum{1}, Gabriel Jeanson\affilnum{1}, Alexander Krawciw\affilnum{2}, Effie Daum\affilnum{1}, Xinyuan Qiao\affilnum{2},\\ Sven Lilge\affilnum{2}, Timothy D Barfoot\affilnum{2}, and François Pomerleau\affilnum{1}}

\affiliation{\affilnum{1}Norlab, Université Laval, Quebec City, Canada\\
\affilnum{2}University of Toronto Robotics Institute, Toronto, Canada}

\corrauth{Matěj Boxan,
Norlab
Université Laval,
Quebec City, Quebec,
G1V~0A6, Canada.}
\email{matej.boxan@norlab.ulaval.ca}

\begin{abstract}
The \acf{FoMo} dataset is a comprehensive multi-season data collection, recorded over the span of one year in a boreal forest.
Featuring a unique combination of on- and off-pavement environments with significant environmental changes, the dataset challenges established odometry and \acs{SLAM} pipelines. 
Some highlights of the data include the accumulation of snow exceeding \qty{1}{\meter}, significant vegetation growth in front of sensors, and operations at the traction limits of the platform.
In total, the \ac{FoMo} dataset includes over \qty{64}{\kilo\meter} of six diverse trajectories, repeated during 12~deployments throughout the year.
The dataset features data from one rotating and one hybrid solid-state lidar, a \acl{FMCW} radar, full-HD images from a stereo camera and a wide lens monocular camera, as well as data from two \acsp{IMU}.
\acl{GT} is calculated by post-processing three \acs{GNSS} receivers mounted on the \ac{UGV} and a static \acs{GNSS} base station.
Additional metadata, such as one measurement per minute from an on-site weather station, camera calibration intrinsics, and vehicle power consumption is available for all sequences.
To highlight the relevance of the dataset, we performed a preliminary evaluation of the robustness of a lidar-inertial, radar-gyro, and a visual-inertial localization and mapping techniques to seasonal changes.
We show that seasonal changes have serious effects on the \rev{localization} capabilities of state-of-the-art methods.
The dataset and development kit are available at \url{https://fomo.norlab.ulaval.ca}.
\end{abstract}


\maketitle
\let\thefootnote\relax\footnotetext{This article has been accepted for publication in the International Journal of Robotics Research. The final published version will be available via DOI~10.1177/02783649261491481.}

\section{Introduction}
\acresetall
Recent years have seen a surge of datasets for autonomous robot navigation.
Traversing urban environments \citep{Burnett2023}, the Scottish Highlands \citep{Gadd2024}, and city parks \citep{Liu2024}, these datasets serve as an essential tool for supporting research on robot autonomy in various conditions.
Indeed, publicly available datasets and related leaderboards can accelerate the exploration of novel methods in odometry, \ac{SLAM}, or place recognition, in a manner similar to what has been seen in computer vision \citep{Deng2009} or \acl{NLP} \citep{Rajpurkar2016}.
The relevance and success of a dataset depend on several factors, including precise time synchronization, spatial calibration, standardized data formats, and the employed sensor modalities.
However, the size and variety of the recorded data are equally important.
Specifically for on-road conditions, achieving data variety is made possible by the vast global fleet of passenger cars, which function as recording devices across the planet \citep{Caesar2020}.
However, data collection on such a large scale is currently not feasible off-road.
Most works only cover a short span of a few days or weeks \citep{Jiang2021, Gadd2024}, as longer recordings pose additional challenges for both the hardware and the personnel.
Therefore, datasets such as the one proposed by \citet{Maturana2018} and \citet{Sivaprakasam2024} focus on what makes navigation in rough terrain difficult: bumps, high pitch and roll angles, collisions, and other factors that are risky in normal driving operations.
Meanwhile, even datasets spanning over one year, such as \citep{Knights2023}, are typically recorded in locations with stable weather and minimal seasonal variation.

\begin{figure*}[!h]
    \centering
    \includegraphics[width=\textwidth]{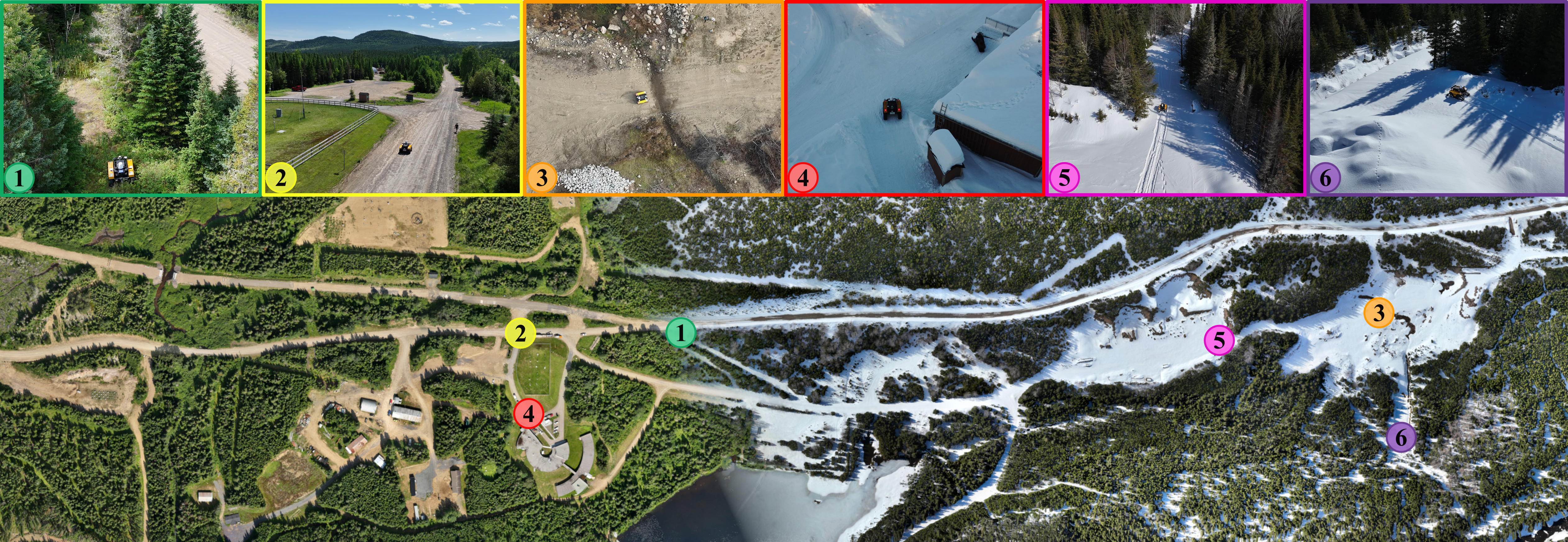}
    \caption{FoMo dataset is a year-long recording capturing seasonal changes of a typical boreal forest.}
    \label{fig:intro}
\end{figure*}

Seasonal changes can play a crucial role in the quality of robot localization and mapping, possibly hindering long-term autonomy in off-road environments \citep{Baril2022, Boxan2025}.
While a dense tree canopy has a higher effect on blocking \ac{GNSS} signals in summer \citep{Ali2020}, visual odometry and \ac{SLAM} methods suffer from illumination changes as traditional \ac{AE} algorithms struggle with snowy and forested environments \citep{Gamache2025}.
Similarly, place-recognition algorithms, often based on camera images and crucial for loop closure detection in \ac{SLAM} systems, struggle with multi-seasonal data \citep{Neubert2013}.

Moreover, \citet{Baril2022} conducted limited field tests, showing that a lidar-inertial odometry system can experience localization failure after a snowstorm due to sudden changes in snow accumulation.
Furthermore, the authors highlighted that terrain traversability evolves from winter to summer months, with their platform becoming immobilized in a mud pit that was originally frozen in winter.

To address these challenges, we present the \ac{FoMo} dataset: a comprehensive, multi-seasonal data collection recorded in a boreal forest, depicted in \autoref{fig:intro}.
Focusing on long-term autonomous navigation, the dataset contains data from two lidars, a camera, a stereoscopic camera, a radar, two \acp{IMU}, and additional sensors.
Spanning over one year and a distance of \qty{64}{\kilo\meter}, the data collection documents the environmental changes in a typical exemplar of a boreal forest in southern Quebec.
At the core of the presented dataset are substantial seasonal changes.
For instance, winter conditions produced snow accumulation exceeding one meter in open areas, with roadside snowbanks often surpassing three meters in height.
To underline the challenges related to the recording location climate, we include detailed measurements from an on-site meteorological weather station covering the full span of the data recording at a rate of one reading per minute.
We provide the dataset in two data formats: a human-readable representation and the ROS2 \texttt{mcap} binary files.
Additionally, we include a public \ac{SDK} for data conversion, calibration, and algorithm evaluation.
Our evaluation focuses on localization and mapping tasks, including odometry and \ac{SLAM}.
The algorithm quality is assessed using three \acp{DOF} \ac{GT} trajectories generated with \ac{PPK} from three \ac{GNSS} receivers and a static base station.
Additionally, the offered \ac{GT} includes per-point uncertainties.
In summary, our main contributions are:

\begin{itemize}
    \item A repeated collection of \qty{6}{\kilo\meter} of trajectories, spanning over one year and totaling \qty{64}{\kilo\meter} that features significant seasonal variations, including high snow accumulation.
    \item A unique combination of both on-road, off-pavement, and off-trail conditions, including the vehicle's high pitch and roll angles, strong vibrations, and sensor occlusion.
    \item A multi-season evaluation of existing lidar, radar, and stereo-inertial localization and mapping methods.
\end{itemize}

\begin{figure}[!b]
    \centering
    \includegraphics[width=1.0\linewidth]{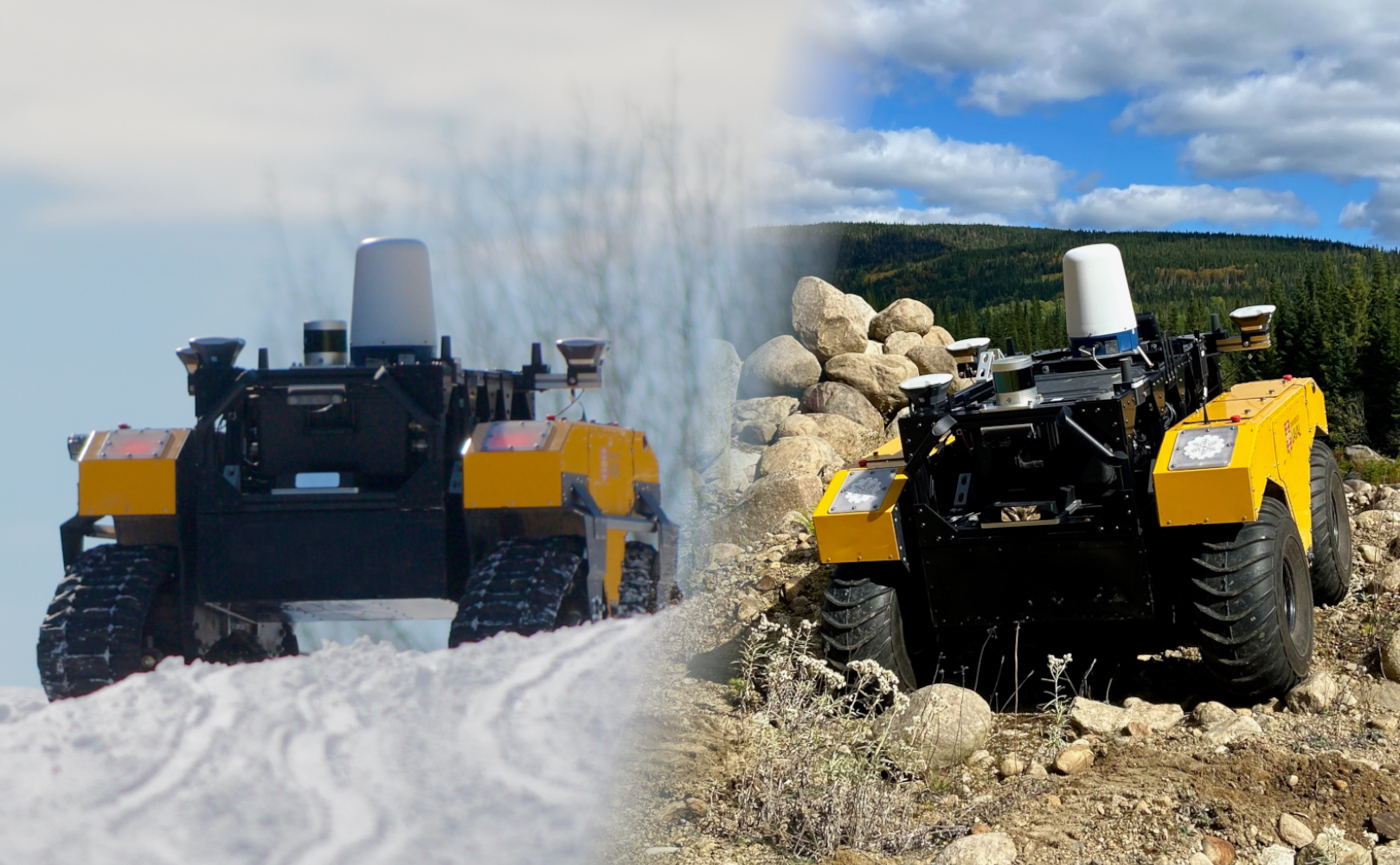}
    \caption{The \ac{UGV} used for the data collection, equipped with tracks on snow (left) and wheels in other seasons (right).}
    \label{fig:data-collection:warthog}
\end{figure}

\section{Related Work}

\begin{table*}[!h]\renewcommand{\arraystretch}{1.15}
	\centering
	\color{black}
	\caption{Overview of the most relevant datasets with deployments spanning multiple seasons.
    We only mention the \acl{GT} related to odometry and localization.
    The label \emph{Off-Pav.} stands for off-pavement.
    }
	\label{tab:sota:overview}
	\footnotesize
	\begin{tabular}{@{}p{3.23cm} p{1.7cm} c c c c c c c c p{3.6cm}@{}}
        \toprule
        \textbf{Name \& References} & \multicolumn{4}{c}{\textbf{Deployment Setting}} & \multicolumn{5}{c}{\textbf{Sensors}} & \textbf{\acl{GT}} \\
        \cmidrule(lr){2-5} \cmidrule(lr){6-10}
                                                          & \centering {\textbf{Environment}}     & \textbf{Distance}&\textbf{Off-Trail}&\textbf{Off-Pav.}& \rot{\textbf{IMU}} & \rot{\textbf{Mono}} & \rot{\textbf{Stereo}} & \rot{\textbf{Lidar}} & \rot{\textbf{Radar}} &  \\
        \midrule
        4Seasons Dataset \newline \citep{Wenzel2021}      & \centering {Urban}                    & \qty{350}{\kilo\meter}  & \Cross     & \Cross     & \Checkmark & \Checkmark & \Checkmark & \Cross     & \Cross     & Pose (GNSS) \\
        Oxford RobotCar \newline \citep{Maddern2017}      & \centering {Urban}                    & \qty{1000}{\kilo\meter} & \Cross     & \Cross     & \Checkmark & \Checkmark & \Checkmark & \Checkmark & \Cross     & Pose (GNSS) \\
		  Boreas Dataset \newline \citep{Burnett2023}       & \centering {Urban}                    & \qty{350}{\kilo\meter}  & \Cross     & \Cross     & \Checkmark & \Checkmark & \Checkmark & \Checkmark & \Checkmark & Pose (GNSS) \\
		FinnForest Dataset \newline \citep{Ali2020}       & \centering {Forest}                   & \qty{33}{\kilo\meter}   & \Cross     & \Checkmark & \Cross     & \Checkmark & \Checkmark & \Cross     & \Cross     & Pose (GNSS) \\
		Goose Dataset \newline \citep{Mortimer2024}       & \centering {Grassland, Woods}         & \qty{162}{\kilo\meter}  & \Cross     & \Checkmark & \Checkmark & \Checkmark & \Cross     & \Checkmark & \Cross     & Pose (D-GNSS)  \\
		MAgro Dataset \newline \citep{Marzoa2024}         & \centering {Plantation}               & \qty{3}{\kilo\meter}    & \Cross     & \Checkmark & \Checkmark & \Checkmark & \Checkmark & \Checkmark & \Cross     & Pose (GNSS) \\
		BLT Dataset \newline \citep{Polvara2024}          & \centering {Plantation}               & \qty{6}{\kilo\meter}    & \Cross     & \Checkmark & \Checkmark & \Checkmark & \Checkmark & \Checkmark & \Cross     & Pose (GNSS) \\
        GrandTour Dataset \newline \citep{Frey2025}       & \centering {Urban, Forest, Mountain}  & N/A                     & \Cross     & \Checkmark & \Checkmark & \Checkmark & \Checkmark & \Checkmark & \Cross     & Pose (GNSS \& \acs{RTS})  \\
        ROVER Dataset \newline \citep{Schmidt2025}        & \centering {Campus, Garden, Park}     & \qty{7}{\kilo\meter}    & \Checkmark & \Checkmark & \Checkmark & \Checkmark & \Checkmark & \Cross     & \Cross     & Position (\acs{RTS}) \\
		Wild-Places \newline \citep{Knights2023}          & \centering {Forest}                   & \qty{33}{\kilo\meter}   & \Checkmark & \Checkmark & \Checkmark & \Checkmark & \Cross     & \Checkmark & \Cross     & Pose (GNSS) \\
		WildScenes \newline \citep{Vidanapathirana2025}   & \centering {Forest}                   & \qty{21}{\kilo\meter}   & \Checkmark & \Checkmark & \Checkmark & \Checkmark & \Cross     & \Checkmark & \Cross     & Pose (GNSS) \\
        \toprule
        \textbf{FoMo (Ours)} & \centering {Forest} & \qty{64}{\kilo\meter} &  \Checkmark & \Checkmark &  \Checkmark & \Checkmark & \Checkmark & \Checkmark & \Checkmark & Position (GNSS) \\
		\toprule
	\end{tabular}
\end{table*}

In this section, we present a selection of notable data collections capturing multi-seasonal data for robotics applications, specifically focusing on autonomous off-road driving.
We further distinguish between off-pavement and off-trail conditions.
While both off-pavement and off-trail could be considered off-road, their difference lies in the complexity of the terrain.
In an off-pavement scenario, a vehicle follows a path that lacks a paved, engineered surface.
These can include gravel roads, forest roads, or firebreaks.
Off-trail, on the other hand, does not feature any path, and the vehicle must therefore navigate through vegetation and around obstacles.
This lack of a clear path makes the off-trail conditions inherently challenging for any autonomous system.
\autoref{fig:data-collection:warthog} provides visual example of these scenarios.
The \ac{UGV} is shown driving off-pavement on compacted snow on the left, and traversing a stone quarry off-trail on the right.

To date, a breadth of mobile robot datasets has been released to facilitate the evaluation of navigation and trajectory estimation algorithms.
However, the majority of currently available data features structured environments, typically with an emphasis on urban settings \citep{Huang2010, Geiger2012}.
Driven by the unique challenges of navigating highly unstructured environments, off-road data recordings have garnered increasing attention to address this gap \citep{Furgale2012,Leung2017,Chebrolu2017,Cheng2021,Triest2022,Liu2024,Gadd2024}.
Currently, both urban and off-road datasets frequently feature idealized weather and lighting conditions.
\rev{Data recordings featuring off-trail terrain and difficult illumination, such as the work of \citet{Gamache2025}, lack repeated traversals of the same trajectories in different seasons.
To address this gap, there is a growing shift toward recording robot operations across multiple months and different seasons \citep{Ali2020,Wenzel2021,Knights2023,Burnett2023}.
 }
We summarize the most relevant datasets in \autoref{tab:sota:overview}, with emphasis on data collections spanning multiple seasons.

Examining the currently available datasets reveals several limitations in the state of the art.
Most notably, some of the most popular multi-season datasets are limited to structured, urban settings \citep{Maddern2017,Wenzel2021,Burnett2023}.
Data collections, featuring off-road driving and targeting \rev{localization}, often lack revisits with longer periods of time between deployments \citep{Jiang2021, Gadd2024}.
Among the datasets concerned with multi-season deployments in off-road environments, we observe that the seasonal changes are either minimal \citep{Knights2023,Vidanapathirana2025}, or the re-deployments are too temporary and sparse re-deployments don't capture the gradual changes of the environment \citep{Ali2020, Baril2022}.

Moreover, many contemporary data collections are constrained by their deployment characteristics.
For instance, agricultural datasets \citep{Marzoa2024, Polvara2024} do not offer the long-duration trajectories critical for navigating highly unstructured environments.
A similar limitation applies to the ROVER dataset \citep{Schmidt2025}, which restricts its coverage to relatively short paths within more structured settings (\ie~campus, garden, and park).
Most datasets are further limited to a subset of the sensors considered in our work.
While the combination of a stereo camera, an \ac{IMU}, and a lidar appears in multiple off-road multi-season data collections \citep{Marzoa2024, Polvara2024, Frey2025}, none of the works include radar.
This missing sensor is a significant limitation, as radars have gained growing interest in recent years for their inherent environmental resilience \citep{Harlow2024}.

The presented \ac{FoMo} dataset aims to address many existing gaps in the current literature.
It features comparatively long trajectories that are repeated multiple times throughout the year, utilizing an exhaustive sensor suite that includes a \ac{FMCW} radar.
Over this period, significant seasonal changes in the boreal forest can be observed, including substantial snow accumulation.
Large portions of the recorded data include trajectories where the robot drives off-trail, which present greater challenges than standard off-pavement scenarios.

\section{Data collection}
\label{sec:data_collection}

\begin{figure}[htbp]
    \centering
    \includegraphics[width=1.0\linewidth]{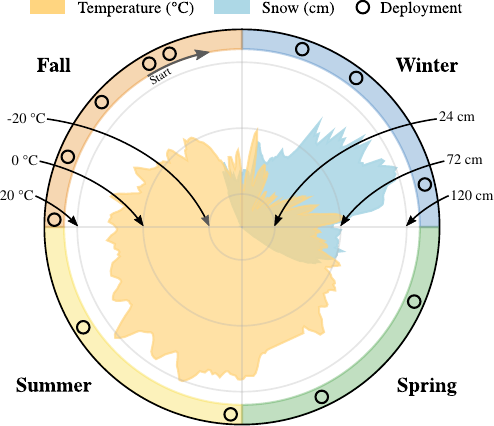}
    \caption{Circular view of daily average temperature and snow depth in meters over a year, from November~2024 to November~2025.}
    \label{fig:data_collection:weather}
\end{figure}

The data collection took place between November 2024 and October 2025 in Forêt Montmorency, a boreal forest located \qty{80}{\kilo\meter} north of Quebec City at an altitude of \qty{600}{\meter}.
Due to its unique location, the site receives approximately \SI{6}{\meter} of snowfall each year \citep{quebec-snow}, which is the highest snowfall in the province of Quebec, leading to snow accumulation of up to \qty{2}{\meter} \citep{Hess2023}.
As a demonstration of the variability of the meteorological conditions this environment offers, \autoref{fig:data_collection:weather} depicts the snow height and ambient air temperature that occurred during data recording.
Notably, the January 29, 2025, deployment had an average temperature of \qty{-19}{\degreeCelsius}, and the March 10, 2025, deployment had over \qty{1}{m} of snow depth.
In contrast, the July 30, 2025, deployment occurred under the warmest conditions, with an average daily temperature of \qty{18}{\degreeCelsius}.
In summary, the presented dataset covers a range of \qty{37}{\degreeCelsius} and features snow on the ground for five months of the year.

\begin{figure*}[htbp]
    \centering
    \includegraphics[width=\textwidth]{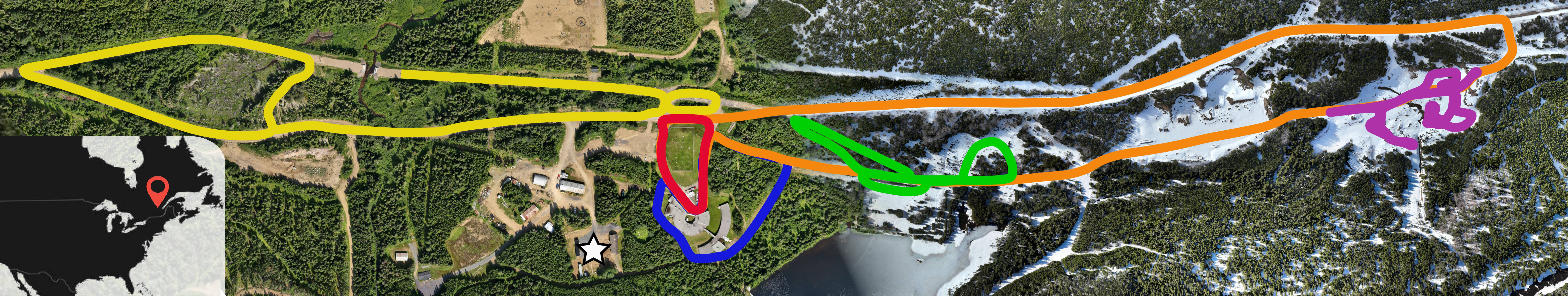}
    \caption{An orthomosaic of Forêt Montmorency, the data recording site, with the six repeated trajectories colored by their label.
    From left to right, we can see \texttt{Yellow}, \texttt{Red}, \texttt{Blue}, \texttt{Orange}, \texttt{Green}, and \texttt{Magenta}.
    \rev{The $\bigstar$ symbol denotes the location of the \acs{GNSS} base station in the area.}
    The lower-left inset shows the location of Forêt Montmorency within North America.
    }
    \label{fig:data_collection:map}
\end{figure*}

The data were collected during both single-day and week-long deployments and were not equally spaced in time.
The six recorded trajectories, color-coded as \texttt{Yellow}, \texttt{Red}, \texttt{Blue}, \texttt{Orange}, \texttt{Green}, and \texttt{Magenta} and displayed in \autoref{fig:data_collection:map}, include various surface types and environmental complexity.
The trajectories were chosen with respect to our data recording platform's mobility and the constraints originating from the chosen \ac{GT} method.
As our \ac{GT} relies on \ac{GNSS}, we preferred wider forest roads for better sky view.
The shortest trajectory, \texttt{Red} (\qty{300}{\meter}), acts as an on-road baseline, confined to paved, semi-urban surfaces.
Two trajectories (\texttt{Green} and \texttt{Magenta}), both \qty{700}{\meter} long, are dedicated to challenging, pure off-trail navigation. 
\texttt{Green} is characterized by steep ascent and descent and dense vegetation that impedes \ac{GNSS} and visual sensing.
\texttt{Magenta}, on the other hand, focuses on the highly irregular, rocky terrain of the stone quarry, including crossing a creek.
The three remaining trajectories feature mixed environments: \texttt{Blue} (\qty{550}{\meter}) and \texttt{Yellow} (\qty{1900}{\meter}) combine on-road with off-pavement gravel road segments, often involving long uphills and significant seasonal challenges like deep snow or snowbanks.
The longest path, \texttt{Orange} (\qty{2200}{\meter}), also mixes on-road and off-pavement.
During dry summer days, dust clouds can be observed on the gravel road that the \ac{UGV} traverses.
The trajectories total 9~intra and 10~inter-sequence revisits.
We provide a complete list of the driven trajectories, along with the deployment day weather, in the \nameref{appendix}.

\begin{figure*}[htbp]
    \centering
    \includegraphics[width=\textwidth]{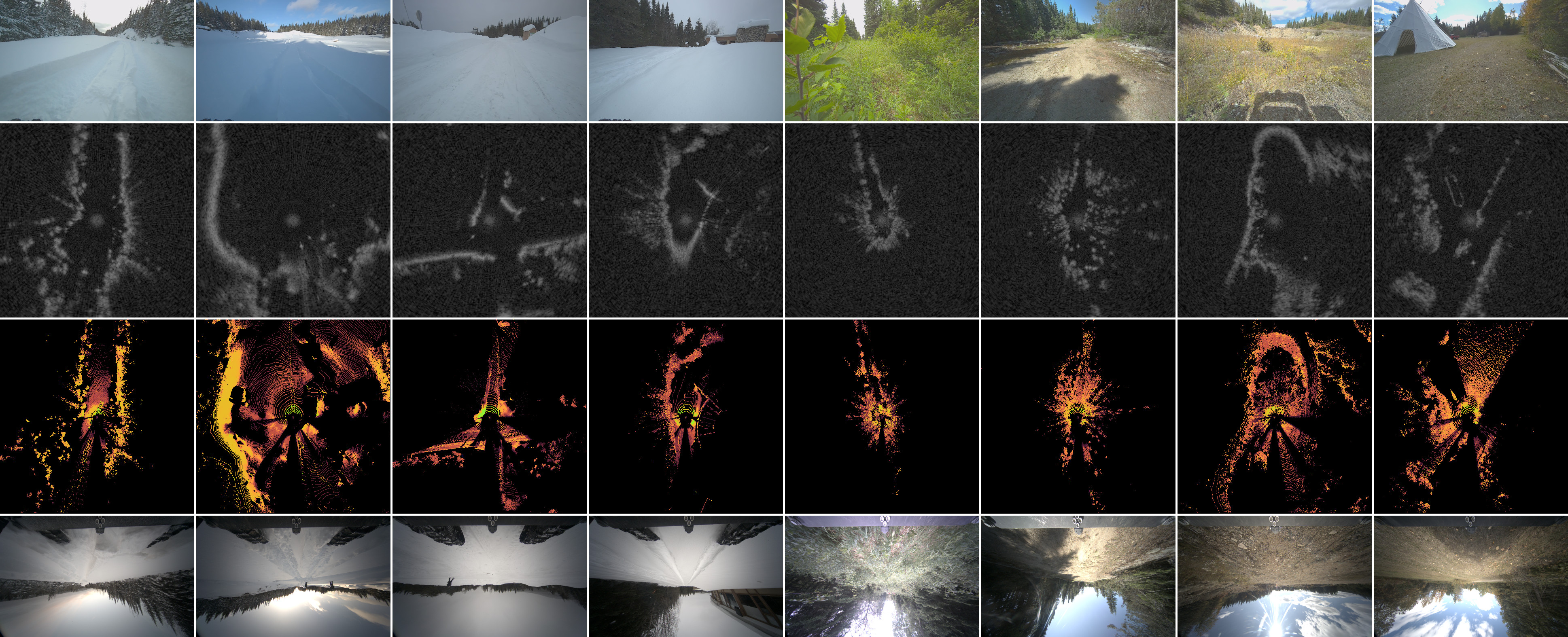}
    \caption{Eight locations of interest captured in three modalities.
    The top row shows images captured from the left stereo camera.
    The second row shows radar scans warped into a Cartesian view.
    The third row shows point clouds from the RoboSense lidar colored by intensity.
    The fourth row shows the rear Basler camera rotated by \qty{180}{\degree}.}
    \label{fig:data_collection:sensor_comparison}
\end{figure*}

\autoref{fig:data_collection:sensor_comparison} provides a visual comparison of four different data modalities across eight diverse locations, including gravel roads, a stone quarry, and densely vegetated trails.
The figure presents camera views (front and rear) in the first and last rows, and radar and lidar scans in the middle two.
Although both scans show similar structures, the 2D radar adds distinct information.
Notably, in the last column, the radar not only reveals the complete structure of the white tent but also demonstrates its wave penetration capability.
A Clearpath Warthog \ac{UGV}, depicted in \autoref{fig:data-collection:warthog}, was utilized for the data recording.
The \ac{UGV} is equipped with a custom-designed sensor frame that hosts two lidars, two cameras, two \acp{IMU}, and an \ac{FMCW} radar sensor.
The frame also accommodates three Emlid Reach M2 receivers with helical antennas used for \ac{GT} calculation.
The platform was mounted on tracks for improved mobility in the winter months, between January and April.
\autoref{fig:sensors:sensors} illustrates the sensors' placement on the robot's frame together with the important coordinate systems, while \autoref{tab:sensors:sensors} details the sensors' specifications.

\begin{table}[htb]
    \centering
    \caption{Sensor specifications ordered by frequencies.}
    \label{tab:sensors:sensors}
    \begin{tabularx}{\columnwidth}{@{}X c r @{}}
    \toprule
        Sensor Type & Model & Recording \\ \midrule
        Wheel encoders                          & 2 $\times$ Hall effect sensors    & \qty{4}{\Hz} \\
        Radar                                   & Navtech CIR-304H                  & \qty{4}{\Hz}\\
        Lidar                                    & RoboSense Ruby Plus              & \qty{10}{\Hz} \\
        Lidar                                    & Leishen LS128S1                  & \qty{10}{\Hz} \\
        Robot GNSS                               & 3 $\times$ Emlid M2              & \qty{10}{\Hz} \\
        Static GNSS                              & Emlid Reach RS3                  & \qty{10}{\Hz} \\
        Stereo camera                           & ZED X                             & \qty{10}{\Hz} \\
        Mono camera                                  & Basler ace2                  & \qty{10}{\Hz} \\
        Pressure sensor                         & Adafruit DPS310                   & \qty{100}{\Hz} \\
        \acs{IMU}                                & Xsens MTi-30                     & \qty{200}{\Hz}  \\
        \acs{IMU}                                & VectorNav VN100                  & \qty{200}{\Hz} \\
        Microphone                              & 2 $\times$    ATR4650             & \qty{16}{\kilo\Hz} \\
    \bottomrule   
    \end{tabularx}
\end{table}

\section{Dataset Format}
We follow the format specification employed by \cite{Burnett2023} in the Boreas dataset.
The \ac{FoMo} dataset is split into \textit{deployments}, which include multiple trajectory \textit{recordings}.
These \textit{deployments} are identified by the data collection day or by the first day of a data collection week as \texttt{YYYY-MM-DD}.
Subsequently, \textit{recordings} are then located within the \textit{deployments} folders, with the format \texttt{name-YYYY-MM-DD-HH-mm}, referring to the name and start of the trajectory.
Each \textit{recording} contains all sensor data, meteorological data, and calibration files.
Additionally, the \textit{recording} folder also includes the \ac{GT} trajectory with per-point covariances.
A file-tree structure of an example \textit{deployment} is provided in the \nameref{appendix}.

\newpage
\subsection{File Description}
The following list contains the specifications of the file format of the recorded sensor measurements.
The \acl{GT} data format is detailed in a later section of this article. 
All timestamps refer to the UNIX time in microseconds.
\begin{itemize}
    \item \textbf{Images:}
    We provide the images from both lenses of the stereo ZED~X camera as rectified \texttt{.png} files.
    Similarly, data from the wide lens Basler were undistorted before saving to \texttt{.png}.
    All image data are provided with a resolution of $1920 \times 1200\text{ px}$.
    The image timestamps correspond to the start of the camera exposure event.
    The images were manually anonymized for faces and license plates using a custom-built pipeline.
    \item \textbf{Lidar scans:}
    The readings from our two lidar sensors are provided as~\texttt{.bin} files with six fields: $[x,~y,~z,~i,~r,~t]$ where $(x,~y,~z)$ are the point's coordinates, $i$ is the intensity of the lidar reflection, $r$ is the ID of the channel which made the measurement and $t$ is the point's timestamp.
    The name of the~\texttt{.bin} matches the timestamp of the first point in each scan.
    We provide Python and Rust scripts to convert between the binary format and~\texttt{.csv} in our \ac{SDK}. Note that the lidar data are not compensated for the \ac{UGV}'s ego motion.
    \item \textbf{Radar scans:}
    For raw radar scans, we follow the Oxford conventions \citep{Barnes2020}, encoding the scans as 2D~polar images of size $M \times (11+R)$, where $M$ is the number of azimuths and $R$ denotes the number of range bins.
    The first eight columns of each image represent each azimuth's timestamp as a 64-bit integer.
    The next two columns represent the rotational encoder value as a 16-bit unsigned integer.
    Finally, the eleventh column is left empty for compatibility with the Oxford format.
    The name of each file corresponds to the timestamp of the beginning of the radar scan.
    Our \ac{SDK} contains a script to convert the radar images between polar and Cartesian coordinates.
    Equally to the lidar data, radar scans are not compensated for the \ac{UGV}'s motion.
    \item \textbf{IMU measurements:}
    Data from the two \acp{IMU} are stored as~\texttt{.csv} files containing $[t,~w_x,~w_y,~w_z,~a_x,~a_y,~a_z]$ where $t$ denotes the timestamp, while $w_x,~w_y,~w_z$ and $a_x,~a_y,~a_z$ denote the angular velocities and the linear accelerations for $(x,~y,~z)$ coordinates, respectively.
    \item \textbf{Audio data:} The data from two microphones are stored independently for the right and left channels in the \texttt{.wav} format, each file containing a \qty{1}{\second} snippet of data.
    \item \textbf{Metadata:}
    Other metadata in the dataset include electric current sensor readings, left and right angular velocities, temperature and pressure measurements, and logs from multiple sensors.
    The full list can be found in the \nameref{appendix}.
    Additionally, we provide the wheel odometry computed using the ideal differential drive model.
    \item \textbf{Climate data:}
    We provide both general weather conditions and snow height information per-trajectory as \texttt{.csv} files.
    These files only cover the time span of a recording; users can download the complete yearly data through the dataset website.
\end{itemize}

\section{Time Synchronization}

\begin{figure}[htbp]
    \centering
    \includegraphics[width=\linewidth]{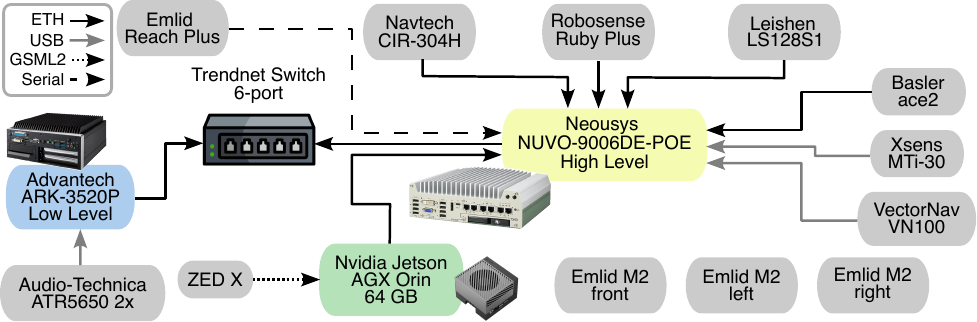}
    \caption{Communication diagram of the FoMo devices.
    All sensors connected via \acf{ETH} are time-synchronized using the IEEE 1588 \acf{PTP}.
    The serial link between Emlid Reach Plus and the Neousys computer is used for \acf{PPS} signal.
    }
    \label{fig:data_collection:comm_diagram}
\end{figure}

Our system, depicted in \autoref{fig:data_collection:comm_diagram}, relies on \acf{PTP} for time synchronization.
Hardware triggering was unavailable due to additional wiring requirements, as well as the limitations of certain sensors.
Our main computer receives a \acf{PPS} signal from an Emlid Reach Plus receiver.
The same computer acts as a \ac{PTP} grandmaster device, synchronizing the internal clock of connected sensors and other computers.
\autoref{fig:data_collection:comm_diagram} provides an overview of the communication protocol employed by each sensor, as well as between the computers.
While the Basler, Leishen, RoboSense, and Navtech all support \ac{PTP} and are connected to the main computer with a direct link, the ZED~X camera uses an automotive \ac{GMSL2} link connected to a secondary Nvidia Jetson Orin AGX computer.
Although the Jetson Orin AGX is time-synchronized through \ac{PTP}, the images from the ZED~X camera are timestamped inside the camera's driver.
To estimate the time delay between the ZED~X shutter time and the timestamp assigned by the camera driver, we generate and capture a stream of matrix barcodes, each embedding absolute UNIX time information $t_{\text{QR}}$.
An offline pipeline then decodes each barcode and calculates the delay by comparing the embedded time with the corresponding driver timestamp $t_{\text{ROS}}$.
The delay is estimated as
\begin{align}
    t_{\text{delay}} = t_{\text{ROS}} - (t_{\text{QR}} + t_{\text{gen}}),
\end{align}
where $t_{\text{gen}}$ is the time in seconds that each barcode's generation requires.
We estimate the mean camera delay $\bar{t}_{\text{delay}}$ over a period of \qty{15}{\minute} as \qty{51.48}{\milli\second} with a standard deviation of \qty{8.78}{\milli\second}.
Additionally, both our \acp{IMU} use USB for connection, depicted in light gray in \autoref{fig:data_collection:comm_diagram}.
Consequently, the \ac{IMU} data are assigned a timestamp when deserialized on the USB interface.
We use \rrev{Kalibr \citep{Furgale2013}} to estimate this latency.
Given the previously estimated mean ZED~X camera delay $\bar{t}_{\text{delay}}$, the delays of the \acp{IMU} is computed as
\begin{align}
    t_{\text{imu}} = \bar{t}_{\text{delay}} - t_{\text{shift}},
\end{align}
giving the latency of \qty{56.09}{\milli\second} for the Vectornav and \qty{60.23}{\milli\second} for the Xsens sensors.
Finally, our low-level computer, while supporting \ac{PTP}, is connected to the main computer through a non-\ac{PTP} switch, which may introduce additional delay between the two devices.
To limit this latency, we minimized network traffic by logging each sensor’s data on the computer to which it was directly connected.

\section{Calibration}

\begin{figure}[h]
    \centering
    \includegraphics[width=1.0\linewidth]{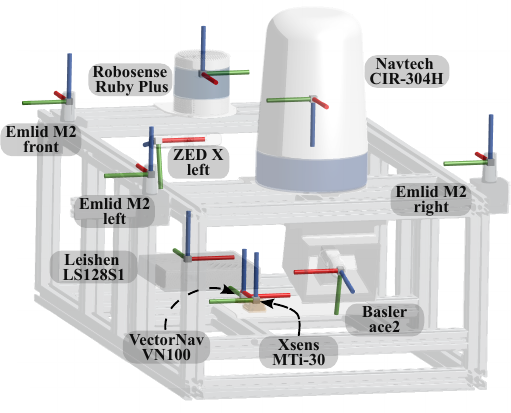}
    \caption{FoMo sensor placement and their coordinate frames.
    The transformation tree is built with the RoboSense at its root. 
    For simplicity, we omit the right lens of the ZED~X camera in this figure.}
    \label{fig:sensors:sensors}
\end{figure}

We performed extensive intrinsic and extrinsic calibrations to find the optimal parameters of our sensor setup.
Indicative placement of the sensor coordinate frames is shown in \autoref{fig:sensors:sensors}.
The resulting calibration files are part of each \textit{recording} directory inside the \texttt{calib} folder.
Additionally, all calibration sequences are available for download on the dataset's website.

\subsection{IMU intrinsics}
We use the Allan variance analysis\footnote{\url{https://github.com/ripl-lab/allan_variance_ros2}} to estimate the \acp{IMU}' noise characteristics and biases.
The recorded sequence used for calculating the noise characteristics of the gyroscope and accelerometer is 24 hours long.
Additionally, we provide the angular velocity biases, computed over a time period of at least \qty{10}{\second}, with the robot being static before data recording started for each trajectory.

\subsection{Cameras Intrinsic}
\label{camera-intrinsic}
To calibrate the intrinsics of the Basler camera, we recorded a sequence with a calibration target measuring \qty{800}{\milli\meter} by \qty{600}{\milli\meter} and featuring $12\times9$ checkers.
Each checker is \qty{60}{\milli\meter} wide.
The recorded images primarily cover the ground behind the robot, with an overfit in this \ac{ROI}, as we envision the primary use of this camera data in terrain assessment tasks.
The intrinsic parameters were then found with the OpenCV camera calibration tool \cite{Bradski2000}.
For the ZED~X camera, we used the intrinsic parameters provided by the camera manufacturer.

\subsection{Lidar-to-Lidar Extrinsics}
\label{lidar-to-lidar-extrinsic}
To find the extrinsic calibration between the RoboSense and the Leishen lidars, we recorded an indoor sequence of lidar scans with a retro-reflective circular target.
This circular target features a reflective ring with inner and outer diameters of \qty{69}{\centi\meter} and \qty{79}{\centi\meter}, respectively.
First, we filtered the reflective ring using per-point intensities.
Subsequently, we fit a set of three circles using \ac{RANSAC} \citep{fischler1981random} to find the target's center point.
After accumulating all fitted centers, we run a point-to-point \ac{ICP} to find the transformation between the two lidars.

\subsection{Lidar-to-Stereo Camera Extrinsics}
\label{lidar-to-stereo-extrinsic}
The extrinsic calibration between the left lens of the ZED X camera and the RoboSense lidar is obtained using a single-shot, targetless calibration method from \cite{Koide2023}.
The calibration scene was recorded in an outdoor parking lot.
We employed the Super-Glue image matching pipeline \citep{Sarling2020} for the initial 2D-3D correspondence search.

\subsection{Lidar-to-Monocular Camera Extrinsics}
As the \ac{FOV} of the rear-facing Basler camera has a minimal overlap with any other sensors, our calibration strategy employs an additional sensor mounted in the back of the \ac{UGV}.
This additional sensor is a Hesai PandarXT-32 lidar.
We construct the RoboSense-Basler transformation as a chain of two transformations.
For the first transformation between the RoboSense and Hesai lidars, we employed the same strategy as in \nameref{lidar-to-lidar-extrinsic}.
The other transformation, between the Hesai lidar and the Basler camera, was then found using the 
single-shot targetless calibration method described in the Section \nameref{lidar-to-stereo-extrinsic}.

\subsection{Lidar-to-Radar Extrinsics}
\begin{figure}[t]
    \centering
    \includegraphics[width=0.99\linewidth]{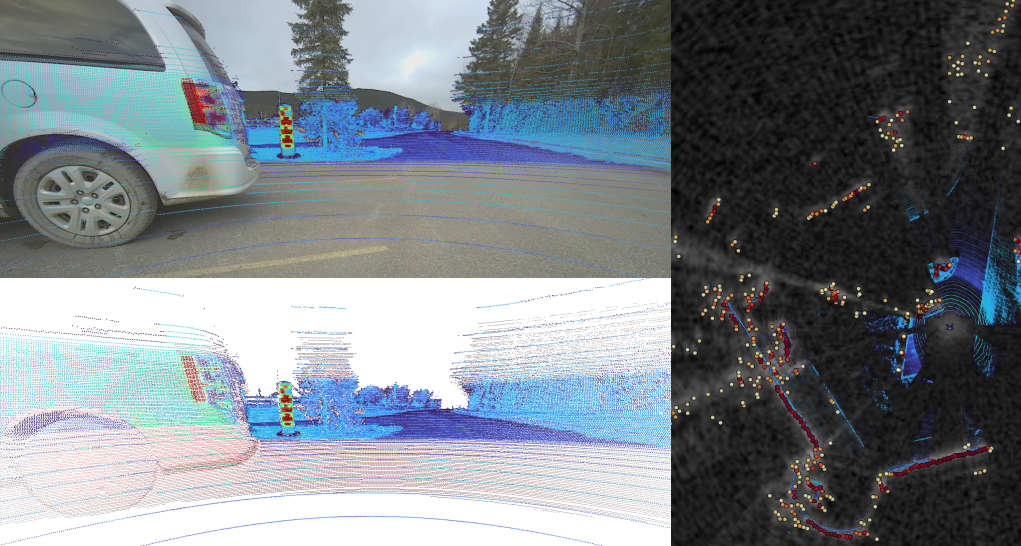}
    \caption{Snippets of sensor alignment results after extrinsic calibration: 
    Top: RoboSense lidar point cloud (colored by intensity) projected onto the ZED~X left camera. 
    Bottom: Leishen lidar points (orange) and RoboSense lidar points (colored by intensity) viewed from the camera perspective.
    Right: \ac{BEV} showing the radar image overlaid with extracted radar points (red, colored by intensity) and RoboSense lidar points (colored by intensity).
    }
    \label{fig:data_collection:extrinsics}
\end{figure}

We estimate the extrinsic transformation between the Navtech radar and RoboSense lidar using a multi-stage \ac{ICP} approach on five static calibration scenes with well-defined geometry.
Initialized from CAD measurements, we optimize for $x$, $y$, and yaw while fixing $z$, roll, and pitch as the Navtech radar is a 2D sensor.
Radar polar images are reconstructed into 2D point clouds.
We start by following the radar image preprocessing steps similar to ~\citep{gentil2025dro}.
Then, point features are extracted using the K-peaks algorithm with a cropped range. K-peaks identifies the top-K strongest contiguous intensity peaks per azimuth with sub-pixel accuracy via range bin averaging.
For lidar, \ac{RANSAC} plane fitting~\citep{fischler1981random} retains only measurements within a vertical band centered about the radar's scanning plane.
Optimization proceeds through five stages from coarse (\qty{25}{\centi\meter} voxels, \qty{1.5}{\meter} threshold) to fine (\qty{3}{\centi\meter} voxels, \qty{15}{\centi\meter} threshold).
The cost function evolves from point-to-point to point-to-plane distances with Tukey robust kernels~\citep{black1996unification}.
Results deviating significantly from CAD priors are rejected as outliers, and the remaining estimates are aggregated via the SE(2) robust median.
The final alignment is shown in \autoref{fig:data_collection:extrinsics}.

\subsection{IMU-to-Camera Extrinsics}
We used the batch optimization toolbox Kalibr \citep{Furgale2013} to estimate the spatial and temporal transformation between the Vectornav and Xsens \acp{IMU} and the left lens of the ZED~X camera.
The camera's frame rate was increased to~\qty{56}{\hertz}, the exposure time fixed to \qty{3}{\milli\second}, and the scene was actively illuminated.
The camera's resolution was set to $1920 \times 1200\text{ px}$ and the checkerboard calibration target previously described in Section \nameref{lidar-to-lidar-extrinsic} was employed.
Additionally, we recorded a validation sequence with lower resolution ($960 \times 600\text{ px}$), together with a sequence replacing the checkerboard with a screen showing an AprilTag target.
As the toolbox requires the perturbation of all 6 \acp{DOF} of the robot with the target being static, we utilized a \qty{900}{\kilo\gram}-capacity crane to lift our platform's sensor frame, weighing over \qty{90}{\kg}, allowing the perturbation of all translational and rotational axes.

\section{Ground Truth}
\label{sec:ground_truth}

Obtaining \acf{GT} trajectories for \ac{SLAM} evaluation poses a significant challenge.
Popular methods such as \ac{MOCAP} systems provide high-speed, millimeter-precision tracking of the platform. 
However, they are restricted to small, confined spaces \citep{Feng2024, Habsi2015}.
Tracking with \acp{RTS}, on the other hand, suffers from the need for a clear line of sight between the station and the \ac{UGV}.
Furthermore, they require stable communication for time synchronization, are time-consuming to deploy, and face challenges when deployed in snow conditions  \citep{Vaidis2023}.
The alternative is to rely on \ac{GNSS} providing absolute positions of the robot in a global coordinate system.
As \ac{GNSS} receivers struggle under tree canopy, we took this into account when planning the trajectories, as described in the Section on \nameref{sec:data_collection}.
To further decrease the uncertainty of the \ac{GNSS} data, we employ \ac{PPK} instead of the traditional \ac{RTK} approach.
Compared to \ac{RTK}, \ac{PPK} eliminates the need for stable communication between the \ac{UGV} and the base antenna, allowing for larger-scale deployments.
Furthermore, \ac{PPK} also allows applying offline corrections with bi-directional filtering, using future observations to help resolve past ambiguities.

To generate our \ac{GT} data, we follow a multi-step pipeline for each trajectory of each deployment.
We post-process the recorded \ac{GNSS} \ac{RINEX} data using \ac{PPK} in Emlid Studio, a software suite based on RTKLib.\footnote{\url{https://github.com/tomojitakasu/RTKLIB}}
\rev{A list of parameters employed for the \ac{GNSS} data processing in provided in the \hyperref[appendix:gnss]{Appendix}.
The location of our base station, an Emlid Reach RS3 receiver, is illustrated in \autoref{fig:data_collection:map}.
The base station was placed in an open-sky plowed area, \qty{1.5}{\kilo\meter} from the farthest point of the recorded trajectories.
We use data from a \ac{CORS}, located about \qty{62}{\kilo\meter} south in Quebec City, to correct the positions of this base station.
}
The corrected reference station's positions are subsequently employed to correct a rover trajectory $\tau^k(t)=\{\bm{p}^k_t\}$ of each of the $k\in\{1,2,3\}$ \ac{GNSS} receivers mounted on the \ac{UGV} where $t$ represents time.
In addition to the points $\bm{p}^k_t$, the postprocessing also provides the estimated covariance matrices $\bm{W}^k_t$.
Next, to combine the three \ac{GNSS} trajectories $\tau^k(t)$ into a single $\hat{\tau}(t)=\{\bm{\hat{p}}_t\}$, we formulate an optimization problem based on the point-to-Gaussian distance metric.
Given a known set of three non-collinear points $\{\bm{q}^1, \bm{q}^2, \bm{q}^3\}$ representing the location of each \ac{GNSS} receiver's antenna on the robot's frame, we compute the error vector
\begin{equation*}
    \bm{e}^k = \bm{p}^k-\bm{\tilde{R}}\bm{q}^k-\bm{\tilde{t}},
\end{equation*}
where $\bm{\tilde{R}}$ and $\bm{\tilde{t}}$ are the prior rotation and translation, and $\bm{p}^k$ is a \ac{GNSS} measurements at a given time.
We then formulate the cost function $J_{p-g}$ as
\begin{equation}
    J_{p-g} = \sum_{k=1}^3 (\bm{e^T}\bm{W^{-1}}\bm{e})_k,
\end{equation}
where $\bm{W^{-1}}$ is the inverse of the covariance matrix $\bm{W}$.
Following \cite{Babin2021}, we decompose the point-to-Gaussian error into three point-to-plane errors, avoiding the need for an iterative solver such as the Gauss-Newton algorithm.
The classic point-to-plane formulation relies on the small-angle approximation.
As the plane defined by the set of the three non-collinear measured points $\{\bm{p}^1_t, \bm{p}^2_t, \bm{p}^3_t\}$ can be in any arbitrary orientation to the plane defined by $\{\bm{q}^1, \bm{q}^2, \bm{q}^3\}$, we initialize the algorithm with an initial transformation matrix $\bm{\check{T}}$.
This transformation matrix is the optimal parameter that minimizes the point-to-point error $J_{p-p}$ between the sets $\{\bm{p}^1_t, \bm{p}^2_t, \bm{p}^3_t\}$ and $\{\bm{q}^1, \bm{q}^2, \bm{q}^3\}$
\begin{equation}
    \bm{\check{T}}=\arg\min_{T}\frac{1}{3}\sum_{k=1}^3(\bm{e^T}\bm{e})_k.
\end{equation}
We use the rotational and translational components $\bm{\check{R}}$ and $\bm{\check{t}}$ of $\bm{\check{T}}$ as the prior rotation and translation for the estimated error vector
\begin{equation*}
    \bm{\check{e}}^k = \bm{p}^k-\bm{\check{R}}\bm{q}^k-\bm{\check{t}}.
\end{equation*}
To find $\bm{\hat{p}}_t$, we compute the point-to-Gaussian distance between the transformed set of points $\{\bm{q}^1, \bm{q}^2, \bm{q}^3\}$ and the set of means and their covariance matrices~$\{\bm{p}^k_t\}$,~$\bm{W}^k_t$~for~$k\in\{1,2,3\}$:
\begin{equation}
    \bm{\hat{T}}=\arg\min_{T}\sum_{k=1}^3(\bm{\check{e}^T}\bm{W^{-1}}\bm{\check{e}})_k.
\end{equation}
The point $\bm{\hat{p}}_t$ is the translation component of the inverse of the optimal parameter $\bm{\hat{T}}$.
Finally, the resulting covariance matrix $\bm{\hat{W_t}}$ is the sample mean of the three $\bm{W}^k_t$. 
The full postprocessing pipeline is summarized in \autoref{alg:gt_generation}.
\rev{Note that based on our observations, RTKLib underestimates the uncertainty in the covariance matrices $\bm{W}^k_t$.
Although the reported uncertainties reflect satellite geometry and suggest millimeter accuracy, the true errors reach centimeter-level in open areas such as forest roads, and tens of centimeters under tree canopy, particularly on the \texttt{Green} trajectory.
}
All \ac{GT} trajectories are provided in the community standard \texttt{tum} format: $(t~x~y~z~qx~qy~qz~qw)$, with $t$ being the timestamps in seconds and $(x,~y,~z)$ are the $\bm{\hat{p}}_t$ point's coordinates.
We substitute $(qx,~qy,~qz,~qw)$ by an identity quaternion, as our method does not provide a reliable orientation estimate.
The covariance matrices for each trajectory, originating from the \ac{PPK} processing, are provided in a separate file.
Note that the timestamps~$t$ do not necessarily correspond to the timestamps of the individual sensor recordings, as the sensors are not triggered by the \ac{GNSS} receivers.
\rev{
Readers can find statistics of the number of observed satellites and the \ac{GNSS} solution type in the \hyperref[tab:gnss_stats]{Appendix}.
Moreover, we provide raw \ac{RINEX} files on our website to support external verification of the \ac{GT} quality.
}

\begin{algorithm}
\caption{\acl{GT} Trajectory Generation}
\label{alg:gt_generation}
\begin{algorithmic}[1]
\State \textbf{Input:} \acsfont{}{RINEX} data from rover and static receivers
\State \textbf{Input:} \acs{CORS} data
\State \textbf{Input:} Antenna positions $\{\bm{q}^1, \bm{q}^2, \bm{q}^3\}$ in robot frame
\State \textbf{Output:} Reference trajectory $\hat{\tau}(t)=\{\hat{\bm{p}}_t,\hat{\bm{W}}_t\}$

\State Correct static ref. station positions using \ac{CORS} data
\State Compute $\tau^k(t)=\{\bm{p}^k_t,\bm{W}^k_t\};~k\in\{1,2,3\}$
\For{each timestep $t$}
    \State \textbf{Initialize transformation:}
    \State Compute $\check{\bm{T}}$ by minimizing point-to-point error:
    \State \quad $\check{\bm{T}} = \arg\min_{T} J_{p-p}(\{\bm{p}^k_t\}, \{\bm{q}^k\})$
    \State \textbf{Optimize using point-to-Gaussian metric:}
    \State Compute error vector: $\bm{e}_k = \bm{p}^k_t - \check{\bm{R}}\bm{q}^k - \check{\bm{t}}$
    \State Compute $\hat{\bm{T}}$ by minimizing point-to-point error:
    \State \quad $\hat{\bm{T}}_t = \arg\min_{T} \frac{1}{3}\sum_{k=1}^3 \bm{e}_k^T (\bm{W}^k_t)^{-1} \bm{e}_k$
    \State Extract fused position: $\hat{\bm{p}}_t = (\hat{\bm{T}}_t^{-1})_{\text{trans}}$
    \State Compute fused covariance: $\hat{\bm{W}}_t = \frac{1}{3}\sum_{k=1}^3 \bm{W}^k_t$
\EndFor

\State \textbf{Return:} $\hat{\tau}=\{\hat{\bm{p}}_t\}$ with $\{\hat{\bm{W}}_t\}$
\end{algorithmic}
\end{algorithm}

\section{Benchmarking Metrics}
\label{sec:benchmark}

Due to the nature of the presented dataset, we particularly focus on odometry and localization tasks in unstructured off-pavement environments.
The goal of our benchmark is to evaluate the robustness of algorithms that employ radar, lidar, camera, or proprioceptive measurements to seasonal changes in our data.
Employing our precomputed \ac{GT} trajectory, we use the same metric as the KITTI dataset \citep{Geiger2012}.
We calculate the translational drift in meters over windows of lengths (\qty{100}{\meter},~\qty{200}{\meter},~$\dots$,~\qty{800}{\meter}) for each estimated trajectory.
Subsequently, we compute the relative drift for each window and report the average value over all windows.
During evaluation, the methods only have access to sensor readings, not the \ac{GNSS} data.

Using a single deployment as the reference, we aim to evaluate the localization capabilities of the tested methods using sensory data from other traversals of the same route.
Note that the proprioceptive approach is an exception in this case, as it is not capable of re-localizing.
However, it still serves as the baseline, showing if the methods' performance degrades significantly when subject to seasonal changes compared to relying on the wheel encoders and \ac{IMU} measurements.
Our evaluation pipeline relies on each method’s ability to construct an environmental representation~$\mathcal{R}_{d_m}=f(\mathcal{S}_{d_m})$ from the sensor data $\mathcal{S}_{d_m}$ of deployment $d_m \in \{1, \dots, N\}$ where N is the total number of traversals of the trajectory.
This representation may take the form of a point cloud, a factor graph, or another suitable structure.
Next, the evaluation procedure initializes the method with the precomputed representation $\mathcal{R}_{d_m}$ and assesses its localization performance using sensor data $\mathcal{S}_{d_l}$ from a different deployment $d_l \in \{1, \dots, N\},~d_l \ne d_m$.
In the next section, we discuss the results of the presented benchmark on four odometry and localization approaches on data from the \texttt{Yellow} trajectory\rev{, recorded on wide forest roads with good satellite visibility.
Consequently, as detailed in the \hyperref[tab:gnss_stats]{Appendix}, the share of \textit{Fixed} \ac{GNSS} solution for this trajectory was never lower than \qty{99}{\percent}.
}

\section{Results}
The primary focus of this section is on evaluating the inter-seasonal robustness of state-of-the-art odometry and localization algorithms.
Additionally, we will discuss the differences in performance between the evaluated methods.
The first tested method is a simple proprioceptive approach integrating wheel velocities and estimating orientation from \ac{IMU} measurements using the Madgwick filter \citep{Madgwick2011}.
This Proprioceptive method serves as a baseline for other methods.
Next, representing lidar-based techniques, we include the lidar-inertial odometry package \texttt{norlab\_icp\_mapper} \citep{Baril2022}.
To test the performance of the \ac{FMCW} radar, we employ the \ac{RGTR} from \cite{Qiao2025}.
Finally, ORB-SLAM3 \citep{Campos2021} uses the stereo ZED~X camera for visual-inertial \ac{SLAM}.

\begin{figure}[tbh]
    \centering
    \includegraphics[width=\linewidth]{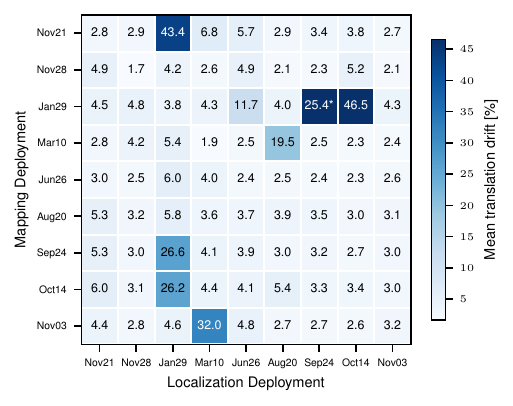}
    \caption{A translation drift matrix for the Lidar-Inertial Odometry method \citep{Baril2022}.
    Each row corresponds to the data used to reconstruct the environment, while each column is the name of the data recording used to localize inside the existing map.
    Symbol~* denotes that the evaluated trajectory covers less than \qty{90}{\percent} of the total trajectory duration as reported by the \acl{GT}.
    }
    \label{fig:eval:confusion-matrix}
\end{figure}

The full results for all deployments for the Lidar-Inertial odometry method \citep{Baril2022} are depicted in \autoref{fig:eval:confusion-matrix}.
In this matrix figure, the values on the diagonal correspond to when the representation $\mathcal{R}_{d_m}$, in this case a 3D~point cloud map, was created for each row.
The measured drift on the diagonal $d_l=d_m$ is in multiple cases the smallest drift on each row (\texttt{Nov28, Jan29, Mar10}).
This result fits with the assumption that a mapping run only accumulates sensor noise from a single deployment.
Other values in the matrix represent \rev{localization} and therefore accumulate additional noise from sensor data from another deployment.
Taking a closer look at the three deployments involving snow (\texttt{Nov28, Jan29, Mar10}), we notice that \texttt{Nov28} does not report notably higher error values than summer months.
Although this late November deployment took place after a snowstorm that brought \qty{40}{\centi\meter} of fresh snow, the drift values are overall lower, staying only \qty{5.0}{\percent} in both \texttt{Nov28} row and column.
\texttt{Jan29} and \texttt{Mar10}, on the other hand, report noticeably higher drift values.
Indeed, the maximum drift value on the \texttt{Jan29} row and column are \qty{46.5}{\percent} and \qty{43.4}{\percent}, respectively.
Similarly, the \texttt{Mar10} row contains a value over \qty{19.5}{\percent}, while the column reaches \qty{32.0}{\percent} in on case.
We attribute this difference between \texttt{Nov28} and the two other months to the presence of snowbanks in the data.
Last but not least, the \texttt{Jan29}-\texttt{Sep24} pair shows a failure of the localization algorithm, denoted with the symbol \textbf{*}.
This failure was caused by the Lidar-Inertial odometry failing to converge to a solution to the \ac{ICP} optimization problem in the last quarter of the trajectory length.
The translation drift matrices for \acl{RGTR} and Stereo-Inertial \ac{SLAM} are presented in the \nameref{appendix}.

\begin{table*}[h]
\renewcommand{\arraystretch}{1.15}
\centering
\caption{Mean translation drifts for four modalities expressed in percentage [\%] along with their standard deviation [$\pm$] using \texttt{Jan29} for initial map representation.
Bold text highlights the best-performing method in each column.
Symbol~\textbf{*} denotes that the evaluated trajectory covers less than \qty{90}{\percent} of the total trajectory duration as reported by the \acl{GT}.
N/A is used when the reported trajectory is too short for the evaluation calculation.
}
\label{tab:eval:comparison}
\footnotesize
\begin{tabularx}{\textwidth}{@{}Xccccccccc@{}}
\toprule
\textbf{Method} & \texttt{Nov21} & \texttt{Nov28} & \texttt{Jan29} & \texttt{Mar10} & \texttt{Jun26} & \texttt{Aug20} & \texttt{Sep24} & \texttt{Oct14} & \texttt{Nov03} \\
\midrule
Proprioceptive \newline \citep{Madgwick2011} & $\mathbf{3.7{}^{}_{\pm1.3}}$ & $4.6{}^{}_{\pm2.1}$ & $15.3{}^{}_{\pm8.1}$ & $4.6{}^{}_{\pm1.8}$ & $\mathbf{3.9{}^{}_{\pm1.4}}$ & $\mathbf{3.4{}^{}_{\pm1.1}}$ & $5.0{}^{}_{\pm2.0}$ & $3.7{}^{}_{\pm1.2}$ & $4.6{}^{}_{\pm1.6}$ \\
\midrule

Lidar-Inertial Odometry \newline \citep{Baril2022} & $4.5{}^{}_{\pm1.5}$ & $4.8{}^{}_{\pm1.7}$ & $3.8{}^{}_{\pm1.5}$ & $\mathbf{4.3{}^{}_{\pm1.8}}$ & $11.7{}^{}_{\pm4.8}$ & $4.0{}^{}_{\pm1.5}$ & $25.4{}^{*}_{\pm13.6}$ & $46.5{}^{}_{\pm28.3}$ & $\mathbf{4.3{}^{}_{\pm1.7}}$ \\
\acl{RGTR} \newline \citep{Qiao2025} & $8.1{}^{}_{\pm3.0}$ & $14.2{}^{}_{\pm6.2}$ & $16.8{}^{}_{\pm5.9}$ & $32.3{}^{}_{\pm11.2}$ & $8.8{}^{}_{\pm3.0}$ & $4.2{}^{}_{\pm1.5}$ & $9.3{}^{}_{\pm4.0}$ & $24.6{}^{}_{\pm10.4}$ & $9.3{}^{}_{\pm3.5}$ \\
Stereo-Inertial \ac{SLAM} \newline \citep{Campos2021} & $1.2{}^{*}_{\pm0.4}$ & $\mathbf{1.5{}^{}_{\pm0.7}}$ & $\mathbf{2.4{}^{}_{\pm1.0}}$ & $7.6{}^{}_{\pm5.4}$ & $8.8{}^{}_{\pm3.8}$ & $1.3{}^{*}_{\pm0.5}$ & $\mathbf{2.3{}^{}_{\pm1.0}}$ & $\mathbf{0.9{}^{}_{\pm0.3}}$ & N/A \\
\bottomrule
\end{tabularx}
\end{table*}

To compare the four methods, we allow each evaluated method to construct an environmental representation $\mathcal{R}_{d_1}=f(\mathcal{S}_{d_1})$ using the data from \texttt{Jan29}.
The results of the comparison are presented in \autoref{tab:eval:comparison}.
Across deployments, the Stereo-Inertial \ac{SLAM} gives the lowest errors in four of nine deployments compared.
We attribute this performance to ORB-SLAM3’s use of loop closure, which the other evaluated methods do not exploit.
However, due to localization errors, the method failed to report the full trajectory three times.
While the method recovered partial results for \texttt{Nov21} and \texttt{Aug20}, the \texttt{Nov03} deployment, recorded at night, proved too difficult for the Stereo-Inertial \ac{SLAM} as the resulting trajectory was too short for the evaluation.
The Proprioceptive method drifts primarily due to wheel slip.
However, this effect is not strictly observed due to the snow. 
\texttt{Nov03} has the second-worst performance but occurs before snowfall, and the \texttt{Mar10} deployment with the deepest snowfall sees the wheel odometry outperform all other methods except the Lidar-Inertial odometry. 
Comparatively, Lidar-Inertial odometry shows low errors in the \texttt{Mar10} and \texttt{Nov03} deployments but struggles to localize in Fall months, as demonstrated by the performance in \texttt{Sept24} and \texttt{Oct24}.
\ac{RGTR} has the largest errors on average. 
\rev{There are multiple reasons why this is true. 
As the \ac{FMCW} radar is a two-dimensional sensor, it struggles to feature-match to the three-dimensional terrain of the \ac{FoMo} dataset.
However, the largest errors appear after sharp turns. 
The low-resolution point clouds produced by the radar for odometry cause it to rotate too far and even using the gyro does not prevent these issues.
Finally, the evaluated translation drift penalizes errors in rotation.
Although other radar-odometry pipelines demonstrated lidar-level accuracy on other datasets, such as \qty{0.61}{\percent} achieved by CFEAR~\citep{Adolfsson2023_CFEAR} on the Boreas dataset, the results are not directly comparable,
}
\rrev{ as the Boreas dataset was collected with an Ackermann-steered vehicle on predominantly flat ground.}

\begin{figure}[htbp]
    \centering
    \includegraphics[width=\linewidth]{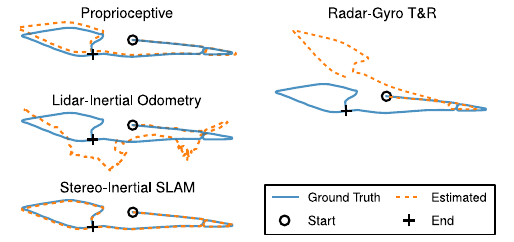}
    \caption{
    Comparison of four methods on the \rev{localization} task inside an environment reconstruction from \texttt{Jan21} using data recorded on \texttt{Oct14}.
    Due to the presence of high snowbanks during the January environment, both Lidar-Inertial Odometry and Radar-Gyro \acs{TaR} suffer high localization errors.
    The proprioceptive sensor does not employ any previous environment reconstruction and serves as a baseline.
    }
    \label{fig:eval:trajectories}
\end{figure}

An example of the effect of a sharp turn on the \acl{RGTR} is displayed in \autoref{fig:eval:trajectories}.
The figure shows the trajectories together with their \acp{GT} for a single environmental representation-localization pair $\mathcal{R}_{d_3},~\mathcal{S}_{d_8}$ (\texttt{Jan29} and \texttt{Oct14}).
The Lidar-Inertial Odometry broke after the opening straight section of the trajectory.
\rev{
The failure occurred in the intersection of two forest roads, both about \qty{10}{\meter} wide, due to a combination of a sharp turn and surrounding snowbanks present in $\mathcal{R}_{d_3}$.
}
Although Radar-Gyro \acs{TaR} showed more robustness, it still incorrectly estimated the turn, drifting away from the original trajectory.
The Stereo-Inertial \ac{SLAM} shows strong performance in this scenario, closing the loop twice and achieving translation drift of \qty{0.9}{\percent}.

\section{Development Kit}
Our \ac{SDK} is a combination of Rust and Python modules that facilitate sensor data fusion, radar processing, or transformation querying.
Additionally, the \ac{SDK} includes all code used for \ac{RINEX} file postprocessing and \ac{GT} generation, as well as the code and configuration used for intrinsic and extrinsic calibrations.
The multi-season evaluation pipeline is based on Docker, and we provide example Dockerfiles for the four tested methods.
We hope these will provide future users with sufficient information to test their own approaches.
As the total size of the dataset exceeds \qty{10}{\tera\byte}, and most users work with a limited set of data modalities, our data download backend enables users to select only the data they are interested in.
Smaller sensor data, such as \ac{IMU}, odometry, \ac{GT} trajectories, and calibration results, are part of every download.
We include scripts to convert between the human-readable data format and ROS2 \texttt{mcap} files.
To make the dataset even more accessible, our server backend supports generating compressed \texttt{mcap} files on demand.
A link to the \ac{SDK} is available at \url{https://fomo.norlab.ulaval.ca}.

\section{Conclusion and Future Work}
In this paper, we presented \acs{FoMo}, a multi-modal dataset for \ac{SLAM} evaluation subject to significant seasonal changes, including snow accumulation exceeding one meter in height.
The dataset features a unique combination of sensors, containing data from two lidars, a stereo camera, as well as a monocular camera, two \acp{IMU}, and an \ac{FMCW} radar.
We discussed the extensive intrinsic and extrinsic calibration of our sensor suit, as well as the time synchronization used to record all sensor data within a common time frame.
All recorded data, including the calibration sequences, are publicly available with a permissive license, along with a public \ac{SDK}.
Additionally, we provide meteorological recordings covering the whole year.

We included preliminary results of a multi-season localization study using three methods that employ diverse sensors, with a Proprioceptive method serving as the baseline.
These results show that further investigations are needed to demonstrate that current localization algorithms are robust to seasonal changes and that our dataset can support such studies.
Moreover, we believe that the \ac{FoMo} dataset value is not limited to odometry and metric localization.
Although we do not offer any specific benchmarks, the community can employ the recorded data as a training dataset for a range of diverse tasks.
These include terrain classification and traversability estimation, especially in the off-trail sequences of \texttt{Magenta} and \texttt{Green} trajectories.
Related to terrain classification is terrain-based power estimation, facilitated by the availability of power data from both the motors and the batteries.
The recorded data include runs featuring both tracks and tires, with current consumption as high as \qty{300}{\ampere}, as well as runs where the robot became immobilized due to excessive snow levels.
Lastly, users in both the off-road autonomous driving and forestry fields may find the dataset valuable for semantic segmentation tasks.
\rev{In the future, we plan on improving the \ac{GT} trajectories and their covariances by utilizing the known geometry of the \ac{GNSS} receivers on the \ac{UGV}, together with the vehicle's known motion model.}
Lastly, the dataset's website will host an evaluation environment, providing an online leaderboard for users aiming to improve the state of the art in \ac{SLAM} algorithms under challenging multi-season conditions.

\begin{acks}
We would like to thank the team at Forêt Montmorency for their assistance in organizing our deployments.
Our gratitude also goes to Vsevolod Hulchuk from the Czech Technical University in Prague, whose efforts helped bootstrap the development of our evaluation pipeline.
Finally, we wish to thank all Norlab members who participated in the data recording sessions.
\rev{The Amazon Open Data Sponsorship program supports this project by hosting the FoMo dataset.}
\end{acks}

\subsection*{\normalsize\sagesf\bfseries Author contributions}
\begin{refsize}\noindent
    This section was included according to IJRR's guidelines.
\end{refsize}

\section*{Statements and declarations}
\subsection*{\normalsize\sagesf\bfseries Ethical considerations}
\begin{refsize}\noindent
    This article does not contain any studies with human or animal participants.
\end{refsize}

\subsection*{\normalsize\sagesf\bfseries Consent to participate}
\begin{refsize}\noindent
    Not applicable
\end{refsize}

\subsection*{\normalsize\sagesf\bfseries Consent for publication}
\begin{refsize}\noindent
    Not applicable
\end{refsize}

\subsection*{\normalsize\sagesf\bfseries Declaration of conflicting interest}
\begin{refsize}\noindent
    The authors declared no potential conflicts of interest with respect to the research, authorship, and/or publication of this article
\end{refsize}

\subsection*{\normalsize\sagesf\bfseries Funding statement}
\begin{refsize}\noindent
This research was supported by the Natural Sciences and Engineering Research Council of Canada (NSERC) and the Fonds de recherche du Québec (FRQNT) through the grant 2023-NOVA-326877 HUNTER (Highlight the Unexpected with Navigation Through Extreme Regions).
Additionally, this project benefited from a Canada Foundation for Innovation Innovation Fund grant (\#39709, PI: E. Thiffault and F. Anctil).
The weather data were acquired from the \emph{Adaptable Earth Observation System} project funded by the Canada Foundation for Innovation, the Government of Quebec, McGill, and UQAM.
\end{refsize}

\subsection*{\normalsize\sagesf\bfseries Data availability}
\begin{refsize}\noindent
    The \ac{FoMo} dataset is available at \url{https://fomo.norlab.ulaval.ca}.
\end{refsize}

\bibliographystyle{SageH}
\bibliography{references}

\appendix
\section*{Appendix}
\label{appendix}

\subsection{Results}
\label{appendix:results}
In this section, we provide complete results for the odometry and \ac{SLAM} methods not detailed in the main text: Proprioceptive method, utilizing the Madgwick filter \citep{Madgwick2011} and wheel odometry, \acf{RGTR} from \citet{Qiao2025}, and Stereo-Inertial \ac{SLAM} from \cite{Campos2021}.
The results for the Proprioceptive method are depicted in \autoref{tab:eval:comparison}, where the Proprioceptive row is identical to the diagonal of a translation drift matrix.
While the majority of reported mean translation drifts lie under \qty{5}{\percent}, the drift during the \texttt{Jan29} recording is over \qty{15}{\percent}.
This high drift was caused by high cover of fresh snow, resulting in the robot being immobilized in the snow multiple times with its tracks spinning freely.
Overall, relying on proprioception only is highly resistant to environmental changes, as long as the \ac{UGV}'s motion model is correctly tuned for both wheels and tracks.

\begin{figure}[h]
    \centering
    \includegraphics[width=\linewidth]{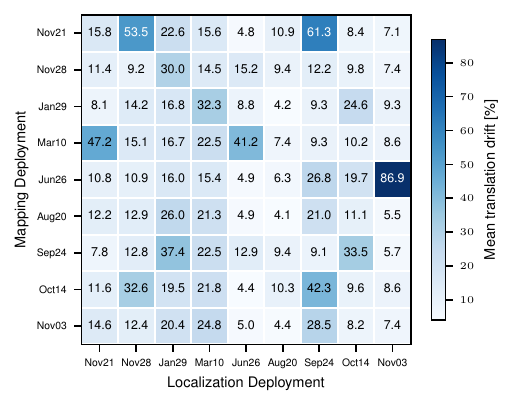}
    \caption{A translation drift matrix for the \acl{RGTR} method \citep{Qiao2025}.
    Each row corresponds to the data used to reconstruct the environment, while each column is the name of the deployment used to localize inside the existing map.
    }
    \label{fig:appendix:confusion-matrix-rtr}
\end{figure}

The full result matrix for the \ac{RGTR} is shown in \autoref{fig:appendix:confusion-matrix-rtr}.
Among all tested methods, \ac{RGTR} reports the highest mean translation drifts.
These errors are typically caused by overestimating the robot's rotation on sharp turns, \rev{as the employed error metric penalizes errors in rotation.}
Although \ac{RGTR} employs a gyro to correct these errors in rotation, creating a globally consistent trajectory in a boreal forest still poses a challenge.
Nevertheless, the method still shows good robustness to seasonal changes when employed in a task employing a local map, such as teach and repeat.
Note that if \ac{RGTR} cannot re-localize, it finishes the trajectory in an odometry mode. 
Therefore, the values reported in the translation drift matrix are a combination of \rev{localization} and odometry errors.
We observe that error accumulates in the z-direction on some runs with large errors. 
This would occur due to numerical instability of the 6-\ac{DOF} optimization \ac{ICP} algorithm used for odometry and localization. 
Constraining the problem such that the 2D points generated by the feature extractor do not allow the state estimate to change in unobserved dimensions.

\begin{figure}[htb]
    \centering
    \includegraphics[width=\linewidth]{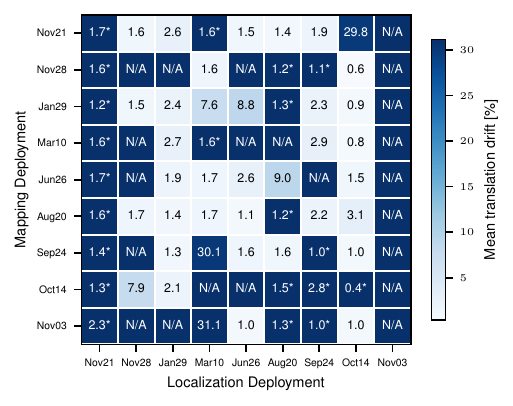}
    \caption{
    A translation drift matrix for ORB-SLAM3 \cite{Campos2021}.
    Each row corresponds to the data used to reconstruct the environment, while each column is the name of the deployment used to localize inside the existing map.
    Symbol~\textbf{*} denotes that the evaluated trajectory covers less than \qty{90}{\percent} of the total trajectory duration as reported by the \acl{GT}.
    N/A is used when the reported trajectory is too short for the evaluation calculation.
    }
    \label{fig:appendix:confusion-matrix-vslam}
\end{figure}

Finally, we report results for the Stereo-Inertial \ac{SLAM} in \autoref{fig:appendix:confusion-matrix-vslam}.
The N/A values in the figure indicate that the final reported trajectory was shorter than the \qty{800}{\meter} segment required for the evaluation, as described in \nameref{sec:benchmark}.
The Stereo-Inertial \ac{SLAM} was the only tested method that utilizes loop closure, which explains the multiple values under \qty{2}{\percent} in the matrix.
Similarly to \ac{RGTR}, if the method cannot re-localize, it switches to odometry and uses place recognition to attempt to recover its position within the previous environment representation.
The Visual \ac{SLAM} generally showed strong performance, and we confirm that loop closure is essential to obtaining highly consistent global maps in boreal forests.
For instance, the multiple values under \qty{1.0}{\percent} in the \texttt{Oct14} column were achieved thanks to repeated loop closures that backpropagated the correction.
However, the method did fail in specific challenging conditions.
The last column (\texttt{Nov03}) was recorded at night with rainfall, where the Visual \ac{SLAM} did not succeed in finding enough matching features in the stereo camera images.
Similarly, in the first column (\texttt{Nov21}), an abrupt rotation by the \ac{UGV} operator broke the Visual \ac{SLAM} localization, causing all output trajectories to cover only the second half of the run.
Whether the observed low error values are due to achieving loop closures or the method's underlying capability to extract meaningful features across seasonal changes is left for future work.

\subsection{File structure and Metadata}
\label{appendix:metadata}
\begin{figure}[!htbp]
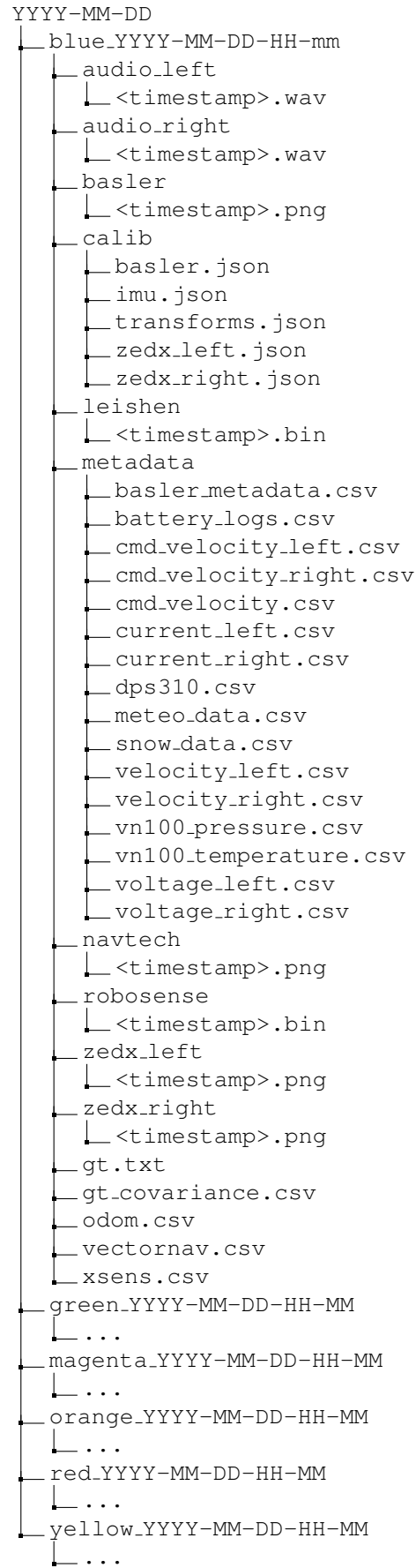

    \dirtree{%
    .1 YYYY-MM-DD.
    .2 blue\_YYYY-MM-DD-HH-mm.
    .3 audio\_left.
    .4 <timestamp>.wav.
    .3 audio\_right.
    .4 <timestamp>.wav.
    .3 basler.
    .4 <timestamp>.png.
    .3 calib.
    .4 basler.json.
    .4 imu.json.
    .4 transforms.json.
    .4 zedx\_left.json.
    .4 zedx\_right.json.
    .3 CHANGELOG.md.
    .3 leishen.
    .4 <timestamp>.bin.
    .3 metadata.
    .4 basler\_metadata.csv.
    .4 battery\_logs.csv.
    .4 cmd\_velocity\_left.csv.
    .4 cmd\_velocity\_right.csv.
    .4 cmd\_velocity.csv.
    .4 current\_left.csv.
    .4 current\_right.csv.
    .4 dps310.csv.
    .4 emlid\_metadata.csv.
    .4 meteo\_data.csv.
    .4 snow\_data.csv.
    .4 velocity\_left.csv.
    .4 velocity\_right.csv.
    .4 vn100\_pressure.csv.
    .4 vn100\_temperature.csv.
    .4 voltage\_left.csv.
    .4 voltage\_right.csv.
    .3 navtech.
    .4 <timestamp>.png.
    .3 robosense.
    .4 <timestamp>.bin.
    .3 zedx\_left.
    .4 <timestamp>.png.
    .3 zedx\_right.
    .4 <timestamp>.png.
    .3 gt.txt.
    .3 gt\_covariance.csv.
    .3 odom.csv.
    .3 vectornav.csv.
    .3 xsens.csv.
    .2 green\_YYYY-MM-DD-HH-MM.
    .3 ....
    .2 magenta\_YYYY-MM-DD-HH-MM.
    .3 ....
    .2 orange\_YYYY-MM-DD-HH-MM.
    .3 ....
    .2 red\_YYYY-MM-DD-HH-MM.
    .3 ....
    .2 yellow\_YYYY-MM-DD-HH-MM.
    .3 ....
    }
    \caption{An example file structure of the \acl{FoMo} dataset \textit{deployment}.}
    \label{fig:dataset_dir_tree}
\end{figure}

An example file structure tree is illustrated in \autoref{fig:dataset_dir_tree}.
Besides the audio files, camera images, encoded radar scans, and lidar point clouds, each \textit{recording} directory, \eg~\texttt{blue-YYYY-MM-DD-HH-mm} inside the \textit{deployment} folder (\texttt{YYYY-MM-DD}) contains calibration files, \acl{GT} with covariances, as well as additional metadata.
The metadata are located inside the \texttt{metadata/} directory.
This directory contains \rev{17} \texttt{.csv} files, as shown in \autoref{fig:dataset_dir_tree}, capturing different aspects of the system's operation: camera information, notably brightness and exposure, power monitoring, motor command velocities, actual motor velocities, reaching \qty{13.25}{\radian\per\second} during the \DTMdate{2025-01-10} deployment on the Orange trajectory, electrical measurements, going as high as \qty{191}{\ampere} during the Magenta trajectory of the \DTMdate{2025-03-10} deployment, environmental sensor data from the DPS310 barometer and Vectornav \ac{IMU}, and weather information, divided in meteorological and snow data.
\rev{Finally, the \texttt{emlid\_metadata.csv} file contains the number of observed satellites, \ac{GNSS} solution and other \acl{GT}-related metadata.}
These data are only available in \texttt{.csv} format and are not exported to \texttt{mcap} rosbags.

\subsection{Trajectories}
\label{appendix:trajectories}

\begin{table*}[t]
    \centering
    \caption{Table of deployments in the FoMo dataset.  
    Trajectory legend: \Checkmark\;Available, \Cross\;Not available, \Alternative\;Alternative route.
    Conditions: \Sun\;Clear sky, \Cloud\;Clouds, \Snow\;Snowfall, \Rain\;Rain, \Night\;Night.}
    \label{tab:deployments}
    \begin{tabularx}{\textwidth}{
        p{2.0cm}  
        >{\centering\arraybackslash}p{1.2cm}  
        >{\centering\arraybackslash}p{1.2cm}  
        >{\centering\arraybackslash}p{1.5cm}  
        >{\centering\arraybackslash}p{1.2cm}  
        >{\centering\arraybackslash}p{1.2cm}  
        >{\centering\arraybackslash}p{1.2cm}  
        >{\centering\arraybackslash}p{1.2cm}  
        >{\centering\arraybackslash}p{2.5cm}  
        }
        \toprule
    \textbf{Date}   & \multicolumn{6}{c}{\textbf{Trajectory}}                                                                   & \textbf{Conditions} & \textbf{Drivetrain}\\
        \cmidrule(lr){2-7}
                             & \textbf{Red} & \textbf{Blue}  & \textbf{Green}& \textbf{Magenta}  & \textbf{Yellow}   & \textbf{Orange}            \\
        \DTMdate{2024-11-21} & \Checkmark   & \Checkmark     & \Checkmark    & \Checkmark        & \Checkmark        & \Checkmark        & \Cloud  & Wheels  \\
        \DTMdate{2024-11-28} & \Checkmark   & \Cross         & \Cross        & \Checkmark        & \Checkmark        & \Cross            & \Sun  & Wheels    \\
        \DTMdate{2025-01-10} & \Checkmark   & \Cross         & \Checkmark    & \Checkmark        & \Cross            & \Checkmark        & \Sun  & Tracks    \\
        \DTMdate{2025-01-29} & \Checkmark   & \Checkmark     & \Alternative  & \Checkmark        & \Checkmark        & \Checkmark        & \Sun  & Tracks    \\
        \DTMdate{2025-03-10} & \Checkmark   & \Checkmark     & \Alternative  & \Checkmark        & \Checkmark        & \Checkmark        & \Snow & Tracks    \\
        \DTMdate{2025-04-15} & \Checkmark   & \Checkmark     & \Cross        & \Cross            & \Cross            & \Cross            & \Rain & Wheels    \\
        \DTMdate{2025-05-28} & \Checkmark   & \Checkmark     & \Checkmark    & \Cross            & \Cross            & \Cross            & \Sun  & Wheels    \\
        \DTMdate{2025-06-26} & \Checkmark   & \Checkmark     & \Checkmark    & \Checkmark        & \Checkmark        & \Checkmark        & \Sun  & Wheels    \\
        \DTMdate{2025-08-20} & \Checkmark   & \Checkmark     & \Checkmark    & \Checkmark        & \Checkmark        & \Checkmark        & \Sun  & Wheels    \\
        \DTMdate{2025-09-24} & \Checkmark   & \Checkmark     & \Checkmark    & \Checkmark        & \Checkmark        & \Checkmark        & \Sun  & Wheels    \\
        \DTMdate{2025-10-14} & \Checkmark   & \Checkmark     & \Checkmark    & \Checkmark        & \Checkmark        & \Checkmark        & \Sun  & Wheels    \\
        \DTMdate{2025-11-03} & \Checkmark   & \Checkmark     & \Checkmark    & \Checkmark        & \Checkmark        & \Checkmark        & \Night\Rain & Wheels     \\
        \bottomrule
    \end{tabularx}
    \vspace{-1mm}
\end{table*}

\begin{figure*}[!h]
    \centering
    \includegraphics[width=\textwidth]{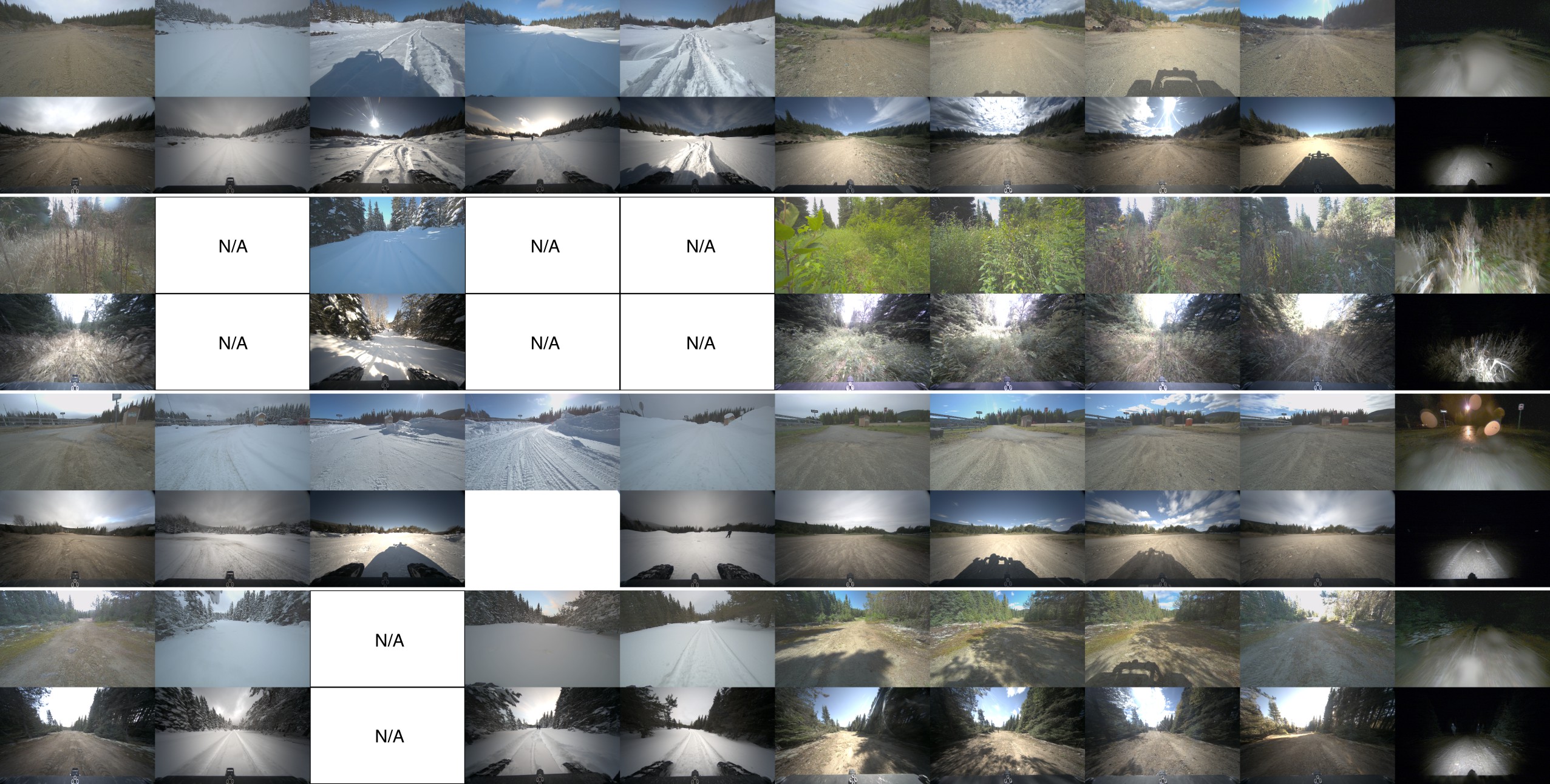} 
    \caption{The forward and backward view from the robot as it drives through key locations across seasons.
    The locations are separated by white horizontal lines.
    Time progresses along the horizontal axis with each image in a row taken in different deployments throughout the year.
    Each column features different locations from the same deployment.
    N/A indicates a missing trajectory in the deployment.
    The rear camera auto-exposure failed in the deployment depicted in row 6 column 4.
    }
    \label{fig:data_collection:season_comparison}
\end{figure*}

\ac{FoMo} dataset was collected over the course of a year, featuring a variety of weather conditions.
\autoref{tab:deployments} summarizes the different trajectories and weather conditions found in each deployment.
Interestingly, the \texttt{Red} baseline trajectory was the only trajectory to be completed in all 12 deployments, as it features a simple on-road loop or just over \qty{300}{\meter}.
During the two deployments featuring the most snow (\DTMdate{2025-01-29}, and \DTMdate{2025-03-10}), an alternative route was taken for the \texttt{Green} trajectory, as the steep terrain of the original path was not traversable as a result of high snow accumulation.
For the \DTMdate{2024-11-28} deployment, \texttt{Blue}, \texttt{Green}, and \texttt{Orange} trajectories were not recorded because of access restrictions during the cross-country skiing season in Forêt Montmorency.
For the \DTMdate{2025-01-10} and \DTMdate{2025-05-28} recordings, the missing trajectories are due to battery issues that prevented us from continuing with the deployment.
During the \DTMdate{2025-04-15} deployment, the robot was back on its wheels after all winter on the tracks.
Hence, melting snow made navigation difficult, leaving only the on-road portions of the trajectories accessible.
Most of the trajectories were recorded during sunny days.
Exceptions include deployment on \DTMdate{2024-11-21} which was cloudy; \DTMdate{2025-03-10}, which saw snowfall; and \DTMdate{2025-04-15}, which was rainy. 
The final deployment, on \DTMdate{2025-11-03}, took place at night during rainfall, with additional lights mounted on the \ac{UGV} to enable camera recording.

\autoref{fig:data_collection:season_comparison} illustrates the views captured by the front ZED~X and the rear Basler cameras from four specific key locations across ten distinct field deployments.
The four reference locations are presented in order from top to bottom: the first is a section of the stone quarry that is crossed by the \texttt{Orange} trajectory; the second is a densely vegetated uphill section forming part of the \texttt{Green} trajectory; the third is an intersection between the \texttt{Red} and \texttt{Orange} trajectories situated near a parking lot, notably featuring high snow banks; and lastly, the fourth location is a woods section found between two roads on the \texttt{Yellow} trajectory.
Time advances along the horizontal axis in the figure, commencing with the first deployment on \DTMdate{2024-11-21} and concluding with the night deployment nearly one year later on \DTMdate{2025-11-03}.
The views from the \DTMdate{2025-04-15} and \DTMdate{2025-05-28} deployments were intentionally omitted, as these particular field sessions did not include the specific trajectories where the reference locations appear.
Data gaps visible within the figure are accounted for by various operational constraints:
The N/A values present on the second pair of rows indicate a missing recording of the \texttt{Green} trajectory.
This specific recording was not made because the slope of that section was deemed too steep to climb, given the prevailing weather conditions during that deployment.
The missing data observed in the last pair of rows resulted from the early termination of the entire deployment due to an unforeseen failure of the robot's batteries.
Finally, the white image displayed in row 6, column 4 is attributed to a failure of the Basler camera's auto-exposure mechanism, a fault which occurred during that particular recording, likely caused by the extreme cold temperature, which was measured below \qty{-20}{\degreeCelsius}.

\rev{
\ac{GNSS} \acl{GT} statistics, including satellite counts and solution types (\textit{Fixed}/\textit{Float}/\textit{Single}), are summarized in \autoref{tab:gnss_stats}.
All values are averaged across the \ac{UGV}'s three receivers per deployment.
The highest share of \textit{Fixed} solutions of all trajectories was recorded on the \texttt{Red} trajectory, with the majority of deployments at \qty{100}{\percent}.
The averaged number of observed satellites varied between 23.9 during the \DTMdate{2025-01-29} deployment and 13.2 in September.
This \DTMdate{2025-09-24} deployment also saw the lowest number of observed satellites for four out of the other five trajectories, with the lowest average of 9.9 observed satellites reported for the \texttt{Green} trajectory.
This trajectory's passage through narrow forest trails led to a minimum \textit{Fixed} solution share among all routes, with \qty{67}{\percent} achieved in September.
The two longest trajectories, \texttt{Yellow} and \texttt{Orange}, benefited from the decision to stay on wide forest roads, whose satellite visibility is not blocked by tree canopy.
The lowest number of observed satellites was therefore at 13.5 and 13.0 for \texttt{Yellow} and \texttt{Orange}, respectively, values similar to the \texttt{Red} trajectory.
While we do not observe any notable difference in \ac{GNSS} solution quality between the winter and other months, we attribute the drop in the number of observed satellites after \DTMdate{2025-06-26} to a firmware update of the Emlid Reach M2 receivers mounted on our robot.
}

\begin{table*}[h]
    \centering
    \caption{
    \rev{
    \ac{GNSS} \acl{GT} Statistics (Solution \% (\textit{Fixed}/\textit{Float}/\textit{Single}) $|$ Average number of observed satellites).
    We omit the \textit{Float} and \textit{Single} percentages when \textit{Fixed} is at \qty{100}{\percent}.
    \Cross~indicates that a trajectory is not available.
    The minimal percentage of \textit{Fixed} solution and the lowest average number of observed satellites in each column are depicted in bold.
    }
    }
    \label{tab:gnss_stats}
    \begin{tabularx}{\textwidth}{p{2.2cm}  >{\centering\arraybackslash}X >{\centering\arraybackslash}X >{\centering\arraybackslash}X >{\centering\arraybackslash}X >{\centering\arraybackslash}X >{\centering\arraybackslash}X}
        \toprule
        \textbf{Date} & \multicolumn{6}{c}{\textbf{Ground Truth Quality (Solution \% $|$ \# Satellites)}} \\
        \cmidrule(lr){2-7}
        & \textbf{Red} & \textbf{Blue} & \textbf{Green} & \textbf{Magenta} & \textbf{Yellow} & \textbf{Orange} \\
        \midrule
        \DTMdate{2024-11-21} & \small \makebox[1.1cm][r]{100} \textbar{} \makebox[0.7cm][l]{19.9} & \small \makebox[1.1cm][r]{100} \textbar{} \makebox[0.7cm][l]{18.0} & \small \makebox[1.1cm][r]{96/3/1} \textbar{} \makebox[0.7cm][l]{13.2} & \small \makebox[1.1cm][r]{99/0/1} \textbar{} \makebox[0.7cm][l]{19.0} & \small \makebox[1.1cm][r]{99/0/0} \textbar{} \makebox[0.7cm][l]{18.8} & \small \makebox[1.1cm][r]{100} \textbar{} \makebox[0.7cm][l]{19.1} \\
        \DTMdate{2024-11-28} & \small \makebox[1.1cm][r]{100} \textbar{} \makebox[0.7cm][l]{23.1} & \small \Cross & \small \Cross & \small \makebox[1.1cm][r]{99/0/0} \textbar{} \makebox[0.7cm][l]{21.2} & \small \makebox[1.1cm][r]{100} \textbar{} \makebox[0.7cm][l]{20.0} & \small \Cross \\
        \DTMdate{2025-01-10} & \small \makebox[1.1cm][r]{100} \textbar{} \makebox[0.7cm][l]{22.1} & \small \Cross & \small \makebox[1.1cm][r]{75/25/0} \textbar{} \makebox[0.7cm][l]{15.3} & \small \makebox[1.1cm][r]{\textbf{67}/33/0} \textbar{} \makebox[0.7cm][l]{20.0} & \small \Cross & \small \makebox[1.1cm][r]{100} \textbar{} \makebox[0.7cm][l]{21.0} \\
        \DTMdate{2025-01-29} & \small \makebox[1.1cm][r]{100} \textbar{} \makebox[0.7cm][l]{23.9} & \small \makebox[1.1cm][r]{\textbf{86}/14/0} \textbar{} \makebox[0.7cm][l]{22.9} & \small \makebox[1.1cm][r]{100} \textbar{} \makebox[0.7cm][l]{21.0} & \small \makebox[1.1cm][r]{80/20/0} \textbar{} \makebox[0.7cm][l]{22.5} & \small \makebox[1.1cm][r]{100} \textbar{} \makebox[0.7cm][l]{18.9} & \small \makebox[1.1cm][r]{91/9/0} \textbar{} \makebox[0.7cm][l]{21.3} \\
        \DTMdate{2025-03-10} & \small \makebox[1.1cm][r]{100} \textbar{} \makebox[0.7cm][l]{21.8} & \small \makebox[1.1cm][r]{99/1/0} \textbar{} \makebox[0.7cm][l]{19.9} & \small \makebox[1.1cm][r]{100} \textbar{} \makebox[0.7cm][l]{23.1} & \small \makebox[1.1cm][r]{100} \textbar{} \makebox[0.7cm][l]{21.4} & \small \makebox[1.1cm][r]{100} \textbar{} \makebox[0.7cm][l]{21.8} & \small \makebox[1.1cm][r]{98/2/0} \textbar{} \makebox[0.7cm][l]{22.5} \\
        \DTMdate{2025-04-16} & \small \makebox[1.1cm][r]{100} \textbar{} \makebox[0.7cm][l]{25.3} & \small \makebox[1.1cm][r]{100} \textbar{} \makebox[0.7cm][l]{22.1} & \small \Cross & \small \Cross & \small \Cross & \small \Cross \\
        \DTMdate{2025-05-28} & \small \makebox[1.1cm][r]{100} \textbar{} \makebox[0.7cm][l]{22.9} & \small \makebox[1.1cm][r]{99/0/0} \textbar{} \makebox[0.7cm][l]{20.7} & \small \makebox[1.1cm][r]{79/20/1} \textbar{} \makebox[0.7cm][l]{17.1} & \small \Cross & \small \Cross & \small \Cross \\
        \DTMdate{2025-06-26} & \small \makebox[1.1cm][r]{100} \textbar{} \makebox[0.7cm][l]{22.3} & \small \makebox[1.1cm][r]{98/1/1} \textbar{} \makebox[0.7cm][l]{18.7} & \small \makebox[1.1cm][r]{97/2/1} \textbar{} \makebox[0.7cm][l]{17.0} & \small \makebox[1.1cm][r]{97/1/2} \textbar{} \makebox[0.7cm][l]{22.2} & \small \makebox[1.1cm][r]{100} \textbar{} \makebox[0.7cm][l]{19.4} & \small \makebox[1.1cm][r]{99/1/0} \textbar{} \makebox[0.7cm][l]{21.0} \\
        \DTMdate{2025-08-20} & \small \makebox[1.1cm][r]{100} \textbar{} \makebox[0.7cm][l]{16.7} & \small \makebox[1.1cm][r]{99/0/0} \textbar{} \makebox[0.7cm][l]{15.0} & \small \makebox[1.1cm][r]{93/6/1} \textbar{} \makebox[0.7cm][l]{12.5} & \small \makebox[1.1cm][r]{99/0/1} \textbar{} \makebox[0.7cm][l]{17.0} & \small \makebox[1.1cm][r]{\textbf{99}/0/1} \textbar{} \makebox[0.7cm][l]{15.9} & \small \makebox[1.1cm][r]{99/0/1} \textbar{} \makebox[0.7cm][l]{16.4} \\
        \DTMdate{2025-09-24} & \small \makebox[1.1cm][r]{100} \textbar{} \makebox[0.7cm][l]{\textbf{13.2}} & \small \makebox[1.1cm][r]{97/3/0} \textbar{} \makebox[0.7cm][l]{\textbf{11.7}} & \small \makebox[1.1cm][r]{\textbf{67}/28/5} \textbar{} \makebox[0.7cm][l]{\textbf{9.9}} & \small \makebox[1.1cm][r]{97/2/1} \textbar{} \makebox[0.7cm][l]{15.2} & \small \makebox[1.1cm][r]{99/1/0} \textbar{} \makebox[0.7cm][l]{\textbf{13.5}} & \small \makebox[1.1cm][r]{96/4/1} \textbar{} \makebox[0.7cm][l]{\textbf{13.0}} \\
        \DTMdate{2025-10-14} & \small \makebox[1.1cm][r]{\textbf{99}/1/0} \textbar{} \makebox[0.7cm][l]{16.7} & \small \makebox[1.1cm][r]{99/1/0} \textbar{} \makebox[0.7cm][l]{13.6} & \small \makebox[1.1cm][r]{96/2/2} \textbar{} \makebox[0.7cm][l]{11.7} & \small \makebox[1.1cm][r]{98/0/1} \textbar{} \makebox[0.7cm][l]{15.3} & \small \makebox[1.1cm][r]{100} \textbar{} \makebox[0.7cm][l]{14.4} & \small \makebox[1.1cm][r]{99/0/1} \textbar{} \makebox[0.7cm][l]{14.6} \\
        \DTMdate{2025-11-03} & \small \makebox[1.1cm][r]{100} \textbar{} \makebox[0.7cm][l]{18.6} & \small \makebox[1.1cm][r]{97/3/1} \textbar{} \makebox[0.7cm][l]{14.4} & \small \makebox[1.1cm][r]{94/5/0} \textbar{} \makebox[0.7cm][l]{12.8} & \small \makebox[1.1cm][r]{100} \textbar{} \makebox[0.7cm][l]{\textbf{15.1}} & \small \makebox[1.1cm][r]{100} \textbar{} \makebox[0.7cm][l]{15.9} & \small \makebox[1.1cm][r]{\textbf{91}/9/0} \textbar{} \makebox[0.7cm][l]{14.5} \\
        \bottomrule
    \end{tabularx}
\end{table*}

\rev{
\subsection{GNSS postprocessing parameters}
}
\label{appendix:gnss}
\rev{
We provide a full list of parameters used for both static and kinematic \ac{GNSS} postprocessing in \autoref{tab:gnss_params}.
These parameters were utilized in Emlid Studio to generate \acl{GT} trajectories, as described in the \nameref{sec:ground_truth} section.
}

\begin{table}[htb]
    \centering
    \caption{
    \rev{Emlid Studio processing parameters used for \acl{PPK} positioning.
    Legend: \acf{SNR}, \acf{IAR} resolves carrier-phase ambiguities.
    GPS, GLONASS, and BDS are the American, Russian, and Chinese global navigation satellite systems.}
    }
    \label{tab:gnss_params}
    \begin{tabularx}{\columnwidth}{
        >{\raggedright\arraybackslash}X
        >{\centering\arraybackslash}p{2.2cm}
    }
        \toprule
        \textbf{Parameter} & \textbf{Value} \\
        \midrule
        \multicolumn{2}{l}{\textit{Static processing}} \\
        \quad Output solution          & One best    \\
        \midrule
        \multicolumn{2}{l}{\textit{Kinematic processing}} \\
        \quad Use base observation     & On          \\
        \midrule
        \multicolumn{2}{l}{\textit{Shared}} \\
        \quad Filter type              & Combined    \\
        \quad Elevation mask           & \qty{15}{\degree}      \\
        \quad SNR mask L1              & \qty{35}{\dBHz}     \\
        \quad SNR mask L2              & \qty{35}{\dBHz}     \\
        \quad Satellites               & All         \\
        \quad IAR GPS                  & Fix \& Hold \\
        \quad IAR GLONASS              & Fix \& Hold \\
        \quad IAR BDS                  & On          \\
        \bottomrule
    \end{tabularx}
\end{table}

\end{document}

%% file: preamble.tex
\usepackage[colorlinks,bookmarksopen,bookmarksnumbered,citecolor=red,urlcolor=red]{hyperref}

\usepackage[english,noabbrev,nameinlink]{cleveref}

\usepackage[printonlyused]{acronym}

\usepackage{siunitx}
\usepackage[all]{nowidow}

\usepackage[dvipsnames]{xcolor}

\usepackage{lipsum}

\usepackage{xspace} 
\newcommand{\ie}{i.e.,\xspace{}}
\newcommand{\eg}{e.g.,\xspace{}}

\usepackage{epstopdf}

\usepackage{import}

\usepackage{booktabs}

\usepackage{tabularx}
\usepackage{multirow, multicol}

\usepackage{amssymb,amsfonts,amsmath,amscd}

\usepackage{bm}

\usepackage{algorithm}
\usepackage{algpseudocode}

\usepackage{dirtree}

\usepackage[english]{datetime2}
\DTMsetdatestyle{ddmmyyyy}

\newcommand{\bbm}{\begin{bmatrix}}
\newcommand{\ebm}{\end{bmatrix}}

\newcommand{\rev}[1]{{#1}}
\newcommand{\rrev}[1]{{#1}}

\acrodef{ICP}{Iterative Closest Point} 
\acrodef{MOCAP}{Motion Capture} 
\acrodef{SLAM}{Simultaneous Localization and Mapping} 
\acrodef{IMU}{Inertial Measurement Unit} 
\acrodef{GT}{Ground Truth} 
\acrodef{FoMo}{For\^{e}t Montmorency} 
\acrodef{SDK}{Software Development Kit} 
\acrodef{DoF}{Degree of Freedom} 
\acrodef{GNSS}{Global Navigation Satellite System} 
\acrodef{UGV}{Uncrewed Ground Vehicle} 
\acrodef{FMCW}{Frequency Modulated Continuous Wave}
\acrodef{PTP}[PTP]{IEEE1588 Precision Time Protocol}
\acrodef{NMEA}[NMEA]{National Marine Electronics Association}
\acrodef{PPK}[PPK]{Post Processed Kinematic}
\acrodef{RTK}[RTK]{Real Time Kinematic}
\acrodef{PPS}[PPS]{pulse-per-second}
\acrodef{DL}[DL]{Deep Learning}
\acrodef{GMSL2}[GMSL2]{Gigabit Multimedia Serial Link 2}
\acrodef{INS}{Inertial Navigation System}
\acrodef{AE}{Auto Exposure}
\acrodef{ROI}{Rectangle of Interest}
\acrodef{RANSAC}{Random Sample Consensus}
\acrodef{RTS}{Robotics Total Station}
\acrodef{BEV}{Bird's-eye view}
\acrodef{ETH}{Ethernet}
\acrodef{RINEX}{Receiver Independent Exchange Format}
\acrodef{CORS}{Continuously Operating Reference Station}
\acrodef{DOF}{Degree of Freedom}
\acrodefplural{DOF}[DOFs]{Degrees of Freedom}
\acrodef{RGTR}[RT\&R]{Radar-Gyro Teach and Repeat}
\acrodef{TaR}[T\&R]{Teach and Repeat}
\acrodef{NLP}{Natural Language Processing}
\acrodef{FOV}{Field of View}
\acrodef{IAR}{Integer Ambiguity Resolution}
\acrodef{SNR}{Signal-to-Noise Ratio}

\usepackage{fontawesome5}
\usepackage[table]{xcolor}

\newcommand{\Checkmark}{\textcolor{Green}{\faCheck}}
\newcommand{\Cross}{\textcolor{Red}{\faTimes}}
\newcommand{\Cloud}{\textcolor{Gray}{\faCloud}}
\newcommand{\Sun}{\textcolor{YellowOrange}{\faSun}}
\newcommand{\Snow}{\textcolor{CornflowerBlue}{\faSnowflake}}
\newcommand{\Alternative}{\textcolor{Gray}{\faDirections}}
\newcommand{\Rain}{\textcolor{Blue}{\faCloudShowersHeavy}}
\newcommand{\Night}{\textcolor{Blue}{\faCloudMoon}}

\makeatletter
\newcommand{\showfont}{%
  Font family: \f@family, %
  Font series: \f@series, %
  Font shape: \f@shape, %
  Font size: \f@size pt%
}
\makeatother

\DeclareSIUnit{\dBHz}{dBHz}